\documentclass[preprint,12pt]{elsarticle}

\usepackage[margin=2.0cm]{geometry}% by courtesy of Mico

\usepackage{amsmath,amssymb,amsfonts}
\usepackage{algorithm,algorithmic}
\usepackage{graphicx}
\usepackage{hyperref}
\hypersetup{hidelinks=true}
\usepackage{textcomp}
\usepackage{multirow}
\usepackage{pifont}
\usepackage{booktabs}
\newcommand{\cmark}{\ding{51}}%
\newcommand{\xmark}{\ding{55}}%
\def\BibTeX{{\rm B\kern-.05em{\sc i\kern-.025em b}\kern-.08em
    T\kern-.1667em\lower.7ex\hbox{E}\kern-.125emX}}

\usepackage{lineno}
\usepackage{xcolor}

\journal{Medical Image Analysis}

\begin{document}

\begin{frontmatter}

\title{CardioMorphNet: Cardiac Motion Prediction Using a Shape-Guided Bayesian Recurrent Deep Network}

\author[1]{Reza Akbari Movahed\corref{cor1}}
\author[1]{Abuzar Rezaee}
\author[2,4]{Arezoo Zakeri}
\author[3]{Colin Berry}
\author[1]{Edmond S. L. Ho}
\author[1]{Ali Gooya}

\affiliation[1]{organization={School of Computing Science, University of Glasgow},
                city={Glasgow},
                country={United Kingdom}}

\affiliation[2]{organization={Division of Informatics, Imaging, and Data Sciences, School of Health Sciences, Faculty of Biology, Medicine and Health, University of Manchester},
                city={Manchester},
                country={United Kingdom}}

\affiliation[4]{organization={Centre for Computational Imaging and Modelling in Medicine (CIMIM), University of Manchester},
                city={Manchester},
                country={United Kingdom}}

\affiliation[3]{organization={School of Cardiovascular and Metabolic Health, University of Glasgow},
city={Glasgow},
country={United Kingdom}}

\cortext[cor1]{Corresponding author: Reza Akbari Movahed}

\begin{abstract}
Accurate cardiac motion estimation from cine cardiac magnetic resonance (CMR) images is vital for assessing cardiac function and detecting its abnormalities. Existing methods often struggle to accurately capture heart motion because they rely on intensity-based image registration similarity losses that may overlook cardiac anatomical regions. To address this, we propose CardioMorphNet, a recurrent Bayesian deep learning framework for 3D cardiac shape-guided deformable registration using short-axis (SAX) CMR images. It employs a recurrent variational autoencoder to model spatio-temporal dependencies across the cardiac cycle, along with two posterior models for bi-ventricular segmentation and motion estimation. The derived loss function from the Bayesian formulation guides the framework to focus on anatomical regions by recursively registering segmentation maps without using intensity-based image registration similarity loss, while leveraging sequential SAX volumes and spatio-temporal features. The Bayesian modelling also enables the computation of uncertainty maps for the estimated motion fields. Validated on the UK Biobank and M\&M datasets by comparing warped mask shapes with ground-truth masks, CardioMorphNet demonstrates superior performance in cardiac motion estimation, outperforming state-of-the-art methods. Uncertainty assessment shows that it also yields lower uncertainty values for estimated motion fields in the cardiac region compared with other probabilistic-based cardiac registration methods, indicating higher confidence in its predictions. In addition, the clinical indices extraction assessment shows that CardioMorphNet estimates the clinical indices more accurately than other approaches.
\end{abstract}


\end{frontmatter}

\section{Introduction}\label{sec1}
Cardiovascular diseases (CVDs) are a group of prevalent diseases contributing to substantial morbidity and mortality rates worldwide \cite{mendis2011global, wilkins2017european}. They encompass a wide range of disorders that negatively affect the heart, blood vessels, and the overall cardiovascular system, such as coronary artery disease, heart failure, stroke, and peripheral arterial disease \cite{mendis2011global}. It is estimated that 17.8 million individuals died from CVDs worldwide in 2017, accounting for 32\% of all global deaths \cite{roth2018global}. Moreover, they impose a significant financial burden on social and healthcare organisations. For instance, the British Heart Foundation reported that the cost of managing patients with CVDs in the United Kingdom is approximately £19 billion annually, which is relatively high for governments, families, and healthcare organisations \cite{bhf-uk-factsheet}. Early and accurate identification of individuals at significant risk of CVDs is critical for improving patients' quality of life and reducing the financial burden on healthcare systems \cite{franco2011challenges}. However, early identification of individuals at high risk of CVDs is a challenging task due to various reasons, such as non-specific symptoms \cite{jurgens2022state}, inherent heterogeneity \cite{koirala2021heterogeneity}, and the asymptomatic nature of CVDs \cite{polonsky2012cvd}.

Some CVDs are related to cardiac motion abnormalities, leading to a shortage of blood supply to the heart muscle. Additionally, specific cardiac motion patterns can serve as biomarkers to identify individuals at high risk of CVDs \cite{opthof2012association}. Early and accurate diagnosis of these abnormalities could save patients' lives, facilitate treatment, and reduce the economic burden \cite{carluccio2000usefulness,cicala2007prevalence}. Cine cardiac magnetic resonance (CMR) images provide dynamic views of the heart, enabling non-invasive diagnosis of cardiac motion abnormalities and comprehensive assessment of cardiac function and morphology \cite{salerno2017recent,bello2019deep,ismail2022cardiac}. There are two types of CMR image sequences: short-axis (SAX) and long-axis (LAX) sequences. SAX sequences consist of a stack of images acquired over the cardiac cycle and perpendicular to the heart's longitudinal axis, covering the region from the base to the apex. These images provide a detailed view of the heart's internal structures, including ventricles, and are crucial for assessing their volumes, heart wall thickness, and myocardial function. LAX cardiac imaging involves acquiring two-dimensional (2D) cine images along the length of the heart, from the base to the apex, typically in planes that include both ventricles and the atria. Cardiologists visually assess these CMR sequences to analyse cardiac motion and identify heart motion abnormalities. However, visual assessment of the CMR data for analysing cardiac motion is subjective, non-reproducible, error-prone, and time-consuming \cite{luijnenburg2010intra}. Therefore, there is an essential need for an accurate, automated approach to assess cardiac motion using the patient's CMR data.

The preliminary step for automated cardiac motion assessment using CMR data is the motion estimation. Most CMR-based motion estimation techniques rely on image registration methods. These approaches include conventional and deep learning-based methods that estimate the spatial transformation between fixed and moving CMR images throughout the cardiac cycle. In these approaches, the estimated spatial transformations are considered the estimation of cardiac motion. Despite their elegant design, these methods do not effectively estimate cardiac motion because they rely on intensity-based image registrtion similarity metrics for registration between the fixed and warped images, thereby capturing various motions, including those irrelevant to heart motion. Consequently, the derived motion is often contaminated by the influence of irrelevant information. To address this limitation, we propose a 3D shape-guided Bayesian recurrent deep learning framework for cardiac motion prediction using SAX volumes over the cardiac cycle, termed CardioMorphNet. This framework comprises a segmentation module for outputting the reference cardiac shape masks, a recurrent variational autoencoder (\textit{RVAE}) network that represents the spatio-temporal dependencies over the cardiac cycle into a latent embedding, and a registration network that estimates motion by extracting features from two sequential SAX volumes and utilising the latent embedding capturing spatio-temporal dependencies. Due to our Bayesian modelling, CardioMorphNet can supervise motion estimation by matching warped shape masks, warped using the estimated motion fields, with the reference masks, without using any intensity-based image registration loss, while leveraging features from sequential SAX volumes and spatio-temporal dependencies. The contributions of this work are:

\begin{itemize}
  \item Proposing a shape-guided recurrent probabilistic SAX volume registration framework for heart motion prediction that can focus on cardiac anatomical regions without reliance on intensity-based image registration similarity metrics, leading to not capturing irrelevant/implausible deformations for cardiac motion.
  \item Exploring a Bayesian-based image registration formulation that enables the registration model to supervise motion field estimation by matching the warped shape masks and the reference ones, which are obtained by a segmentation network, while it predicts motion field by extracting features from SAX volumes and recurrent spatio-temporal dependencies of the SAX sequence.
  \item Using explicit Bayesian modelling, we derive uncertainty maps of estimated 3D motion fields without relying on empirical approaches, enabling efficient voxel-level confidence analysis of cardiac motion estimation.
  \item Through comprehensive experiments on the CMR UK Biobank and M\&M dataset, we demonstrate that our framework significantly improves cardiac motion estimation over state-of-the-art registration methods by registering cardiac shape masks more precisely, more accurately extracting cardiac clinical indices, and estimating deformation fields with lower uncertainty within cardiac anatomical regions.
\end{itemize}

This paper is organised as follows: Section \ref{sec2} provides the literature review for cardiac motion estimation. The proposed framework and its mathematical foundation are presented in Section \ref{sec3}. Quantitative and qualitative results are given in Section \ref{sec4}. Finally, we discuss our study and provide the conclusion in Sections \ref{sec5} and \ref{sec6}, respectively.

\section{Related Works}\label{sec2}
\subsection{Conventional image registration methods}
Conventional image registration methods rely on iterative algorithms to optimise the alignment of two or more images. Most conventional image-based cardiac motion estimation methods are applied to CMR myocardial-tagged images. These images have several markers on the myocardial tissue that deform with the heart's motion, and these methods can track them during the cardiac cycle. For instance, Osman et al. employed the harmonic phase (HARP) technique to track markers in the tagged CMR images \cite{osman1999cardiac}. Moreover, Chen et al. proposed a three-step approach to track cardiac motion using Gabor filters and the robust point matching (RPM) algorithm \cite{chen2009automated}. They used Gabor filters to locate tag intersections in the frames of tagged CMR images, then utilised the RPM technique to sparsely track myocardial motion. In another work, Liu et al. introduced an incompressible motion estimation algorithm (IDEA) to build a dense representation of the 3D field from CMR-tagged images \cite{liu2011incompressible}. In this work, the HARP method processes the tagged images at each step to obtain the displacement vector fields (DVFs) at each voxel. Next, the IDEA algorithm is used to interpolate velocity fields at intermediate discrete times to obtain a dense estimation of the 3D fields from CMR-tagged images.

In addition to using tagged CMR images, various methods employ geometrical and biomechanical modelling to derive DVFs from dynamic myocardial contours or surfaces \cite{wang2001fast, papademetris2001estimation}. For example, Papademetris et al. introduced a Bayesian framework for myocardial motion estimation using 3D ultrasound images \cite{papademetris2001estimation}. In this framework, the images are first segmented interactively, and then the initial displacements are computed using a shape-tracking algorithm. Next, a dense motion field is calculated using a transversely isotropic linear elastic model. Rueckert et al. proposed a non-rigid image registration method using a free-form deformation (FFD) based on B-splines to estimate the local breast motion in magnetic resonance imaging (MRI) breast images \cite{rueckert1999nonrigid}. In 2012, the Temporal Diffeomorphic Free Form Deformation (TDFFD) method was developed, based on the FFD definition, to quantify motion and strain from sequences of 3D ultrasound volumes \cite{de2012temporal}. This method utilises spatio-temporal B-spline kernels to compute a 4D velocity field, thereby achieving temporal consistency in motion recovery. This technique has been widely used for heart motion estimation in tagged and cine CMR images and other cardiac imaging modalities \cite{shen2005consistent, chandrashekara2004analysis, shi2012comprehensive, tobon2013benchmarking}.

Thirion used the concept of diffusion theory for image-to-image matching and cardiac motion tracking as a conventional cardiac motion estimation method \cite{thirion1998image}. Following this method, Vercauteren et al. proposed a nonparametric diffeomorphic image registration approach based on diffusion theory \cite{vercauteren2007non}. To improve this work, McLeod et al. presented an elastic regularisation term to enhance the invertibility of the deformation transformation \cite{mcleod2011incompressible}. Despite numerous conventional image registration techniques, their use is limited by low accuracy in complex scenarios and substantial computational load. This is especially pronounced in medical image registration tasks, where rapid and precise methods are essential.

\subsection{Deep learning-based image registration methods}
With the rapid advances in deep learning and big data, numerous studies have proposed medical image registration techniques that effectively address the substantial computational load and limited accuracy of conventional methods \cite{chen2021deep}. These methods can be categorised into supervised and semi-supervised approaches. Supervised techniques require the ground-truth DVFs to learn how to align two or more images. The ground-truth DVFs are often obtained through random transformation generation \cite{salehi2018real, eppenhof2018pulmonary, eppenhof2018deformable}, conventional registration methods \cite{sentker2018gdl, fan2019birnet}, and model-based DVF generation techniques \cite{uzunova2017training, sokooti20193d}. Methods that use random transformations for ground-truth DVFs generation suffer from bias in the training procedure and performance degradation, as these ground truths differ primarily from real motion. Although model-based DVF generation techniques yield better image registration performance, they still suffer from bias in their results. Similarly, deep-learning-based methods that rely on conventional methods to generate ground-truth DVFs are limited in both accuracy and computational efficiency. The key to improving the performance of supervised methods is to utilise datasets with known ground-truth DVFs. Nonetheless, the lack of datasets with known ground-truth DVFs limits the accuracy and precision of supervised registration techniques. Uzunova et al. presented a data augmentation approach to enable the training of a deep learning image registration model for cardiac images with limited training data and ground-truth DVFs \cite{uzunova2017training}. This study uses the FlowNet architecture \cite{dosovitskiy2015flownet}, a Convolutional Neural Network (CNN)-based model, to estimate optical flow in cardiac images. Although this method is innovative in enabling training with a limited dataset, it is constrained in generating a diverse set of training pairs of medical images with known ground truths for DVFs.

Due to these restrictions, most studies use semi-supervised approaches for medical image registration and cardiac motion estimation. These methods commonly estimate the DVFs from the fixed and moving images and are trained based on intensity-based similarity between the fixed and moved images. One of the most common deep learning semi-supervised techniques is VoxelMorph \cite{balakrishnan2019voxelmorph}, which derives DVFs using a U-Net CNN model that learns features from a concatenation of moving and fixed images. It is trained based on the intensity-based similarity between the fixed and moved images, as well as the similarity of the anatomical regions in both images, provided that anatomical masks are available. Regarding accuracy and computational load, VoxelMorph outperforms non-learning-based image registration techniques, such as methods in \cite{avants2008symmetric,avants2011reproducible} and NiftyReg approaches \cite{modat2014global,rueckert1999nonrigid, modat2010fast}. It has also been used in several studies to estimate cardiac motion. For instance, Upendra et al. presented a VoxelMorph-based model to estimate LV motion from SAX CMR images \cite{upendra2021cnn}. De Vos et al. integrated a CNN regression model with affine registration to estimate DVFs for fixed and moving CMR images in another study \cite{de2019deep}. Meng et al. used multiple hybrid 2D/3D CNN models to estimate cardiac motion given multi-view CMR images in SAX and LAX planes \cite{meng2022mulvimotion}. They also utilised a shape regularisation module to guarantee consistency in the estimated motion fields in the cardiac region. Their results demonstrate that their method outperforms other methods, such as VoxelMorph, FFD, and the method in \cite{vercauteren2007non}. They also extended this work to propagate a high-resolution LVmyo template mesh to the subject space \cite{meng2023deepmesh}. Regarding the modelling of temporal features in the cardiac sequence, Wu et al. proposed the temporal latent residual network (TLRN), which models temporal dependencies in the latent deformation spaces using sequential residual blocks \cite{wu2024tlrn}.

In terms of variational models, Krebs et al. presented a probabilistic model for CMR diffeomorphic registration \cite{krebs2019learning}. This model learns a low-dimensional probabilistic latent space of deformations using moving and fixed CMR images. Krebs et al. also developed another probabilistic CMR image registration framework to model temporal dependencies across the cardiac sequence to enhance motion prediction \cite{krebs2019probabilistic}. They model these dependencies using temporal convolutional networks in the latent deformation space and employ temporal dropout sampling to enforce the framework to search for temporal dependencies. Moreover, Zakeri et al. introduced a recurrent variational CMR image registration model, DragNet, for registering LAX CMR images and generating synthetic heart motion sequences in the LAX view \cite{zakeri2023dragnet}. Compared with conventional methods, DragNet can register unseen image sequences and generate realistic synthetic LAX images from a single frame. It outperforms VoxelMorph \cite{balakrishnan2019voxelmorph} by learning a prior distribution over spatio-temporal dependencies. Although these approaches improve cardiac motion estimation, they suffer from capturing irrelevant DVFs related to cardiac motion because they rely on intensity-based image registration similarity metrics between the fixed and moved images for the learning registration task.

To mitigate limitations of intensity-based image registration metrics, Qiu et al. utilised a differentiable variant of the mutual information metric for image registration in brain and cardiac images using a B-spline deep learning registration model \cite{qiu2021learning}. To further suppress the restrictions of intensity-based registration techniques, focus on anatomical regions, and enhance the segmentation and registration performances in medical imaging, several studies have proposed joint segmentation-registration methods that concurrently perform segmentation and registration tasks. For example, Qin et al. presented a dual-branch CNN for joint motion estimation and segmentation, named JMS, where DVFs are estimated from cardiac images to warp anatomical masks and train the segmentation branch using a semi-supervised approach \cite{qin2018joint}. Wei et al. also proposed a temporal-consistent segmentation framework (TCSF) for cardiac regions across a sequence of 2D echocardiographic images, with co-learning of segmentation and registration \cite{wei2020temporal}. Their approach comprises two stages. First, the segmentation component is trained using ground-truth masks, while the registration network is optimised using bidirectionally warped images across the sequence. Subsequently, the segmentation network is fine-tuned using pseudo-ground-truth masks obtained by warping the ground-truth masks via the registration network. During this stage, the registration network is further fine-tuned using warped masks at end-diastolic (ED) and end-systolic (ES) frames. Bi et al. introduced a variational multi-branch joint segmentation-registration model, SegMorph \cite{bi2024segmorph}, comprising several components that simultaneously handle biventricular segmentation and motion estimation, and learn a prior distribution for spatio-temporal dependencies. In this work, DVFs are estimated from the fixed and moving SAX images, and ground-truth mask labels from the ED phase are warped to other time points using the DVFs to supervise the segmentation branch.

Although these joint segmentation-registration methods enrich cardiac segmentation by generating pseudo-ground-truth masks for their segmentation module, the resulting DVFs lack cardiac focus because their registration branch is trained based on an intensity-based similarity loss between fixed and moved images, potentially leading them to overlook cardiac anatomical regions. In these methods, the segmentation branch benefits only from the warped masks for semi-supervised training, thereby improving segmentation performance. However, it cannot effectively guide the registration branch to focus on the cardiac regions.

\section{Materials and Methods}\label{sec3}
\subsection{Proposed Bayesian Modelling}
Our objective is to estimate cardiac motion in the LV, RV, and LVmyo regions using Bayesian modelling based on an SAX sequence, focusing only on these relevant anatomical areas. Let $\mathcal{I}\!\!= \!\!\{\textbf{I}_{t}\}_{t=0}^{T-1}$, $\mathcal{M}\!\!=\!\! \{\textbf{M}_{t}\}_{t=0}^{T-1}$, $\mathcal{D}\! = \!\!\{\textbf{D}_{t}\}_{t=0}^{T-1}$, and $\mathcal{Z}\!\!=\!\! \{\textbf{Z}_{t}\}_{t=0}^{T-1}$ be the set of SAX volumes, cardiac anatomical masks, DVFs, and latent variables over the cardiac cycle with \textit{T} number of time steps. DVFs are considered estimates of cardiac motion, and the latent variables aim to capture the distribution of spatio-temporal dependencies among SAX volumes throughout the cardiac cycle. For the mentioned sets, $\textbf{I}_{t} \!\in \!\mathbb{R}^{1 \times H \times W \times D}$, $\textbf{M}_{t}\! \in \!\mathbb{R}^{K \times H \times W \times D}$, $\textbf{D}_{t} \!\in \!\mathbb{R}^{3 \times H \times W \times D}$, and  $\textbf{Z}_{t} \!\in \!\mathbb{R}^{1 \times \frac{H}{16} \times \frac{W}{16} \times D}$ are the SAX volume, cardiac anatomical mask, DVF, and latent variable at the \textit{t} time step, respectively. \textit{H}, \textit{W} and \textit{D} denote the SAX volumes' height, width, and depth, and \textit{K} is the number of classes in the cardiac masks. In this study, \textit{K} is set to 4, which represents the background, LV, LVmyo, and RV regions. We denote by $t_{\mathrm{ED}}$ and $t_{\mathrm{ES}}$ the ED and ES time points, respectively, for which ground-truth anatomical masks are available. Given $\mathcal{I}$ and ground-truth anatomical masks for $t_{\mathrm{ED}}$ and $t_{\mathrm{ES}}$, $\textbf{M}_{t_\mathrm{ED}}$ and $\textbf{M}_{t_\mathrm{ES}}$, we aim to estimate $\mathcal{D}$, $\mathcal{Z}$, and $\mathcal{M}' = \{\textbf{M}_{r}| 0\leq r \leq T-1, r \notin \{t_{ED}, t_{ES}\}\}$ that is the set of the anatomical masks at time steps other than $t_{\mathrm{ED}}$ and $t_{\mathrm{ES}}$. This is achieved by estimating the posterior function, $\mathcal{Q} = q(\mathcal{M}', \mathcal{D}, \mathcal{Z}| \mathcal{I}, \textbf{M}_{t_\mathrm{ED}}, \textbf{M}_{t_\mathrm{ES}})$, and learning the joint probability, $\mathcal{P} = p(\mathcal{M}, \mathcal{I}, \mathcal{D}, \mathcal{Z})$. The joint probability, $\mathcal{P}$, is defined using the chain rule as follows:
\begin{equation}\label{eq1}
    \begin{split}
        \mathcal{P} = p(\mathcal{M}, \mathcal{I}, \mathcal{D}, \mathcal{Z}) = \prod_{t=0}^{T-1} &p(\textbf{M}_{t}| \textbf{M}_{< t}, \textbf{I}_{\leq t}, \textbf{D}_{\leq t}, \textbf{Z}_{\leq t}) \\
        & p(\textbf{I}_{t}| \textbf{M}_{< t}, \textbf{I}_{< t}, \textbf{D}_{\leq t}, \textbf{Z}_{\leq t}) \\
        & p(\textbf{D}_{t}| \textbf{M}_{< t}, \textbf{I}_{< t}, \textbf{D}_{< t}, \textbf{Z}_{\leq t}) \\
        & p(\textbf{Z}_{t}| \textbf{M}_{< t}, \textbf{I}_{< t}, \textbf{D}_{< t}, \textbf{Z}_{< t}). \\
    \end{split}
\end{equation}
Similarly, the posterior distribution, $\mathcal{Q}$, for estimating the mask shapes, latent variables, and DVFs, is factorised by the chain rule as follows:
\begin{equation}\label{eq2}
    \begin{split}
        \mathcal{Q} = \!\!\!\!\!\!\!\!\!\!\prod_{\substack{r=0 \\ r \notin \{t_{ED}, t_{ES}\}}}^{T-1} \prod_{t=0}^{T-1} & q(\textbf{M}_{r}| \textbf{M}_{< t}, \textbf{D}_{\leq t}, \textbf{Z}_{\leq t}, \textbf{I}_{\leq t}) \\
        & q(\textbf{D}_{t}| \textbf{D}_{< t}, \textbf{M}_{< t}, \textbf{Z}_{\leq t}, \textbf{I}_{\leq t}) \\
        & q(\textbf{Z}_{t}| \textbf{Z}_{< t}, \textbf{D}_{< t}, \textbf{M}_{< t}, \textbf{I}_{\leq t}). \\
    \end{split}
\end{equation}
It is important to note that cardiac shape masks at $t_{\mathrm{ED}}$ and $t_{\mathrm{ES}}$, $\textbf{M}_{t_{\mathrm{ED}}}$ and $\textbf{M}_{t_{\mathrm{ES}}}$, are observed in this study. Hence, these masks are omitted in the posterior function for estimation. To make our Bayesian modelling computationally tractable and to express the dependencies in a form that can be implemented using neural network components, it is necessary to simplify the joint and posterior functions expressed in Eqs. \eqref{eq1} and \eqref{eq2}, respectively.

\begin{figure}[t!]
  \centering
    \resizebox{0.40\textwidth}{!}{
    \includegraphics[width=0.4\textwidth]{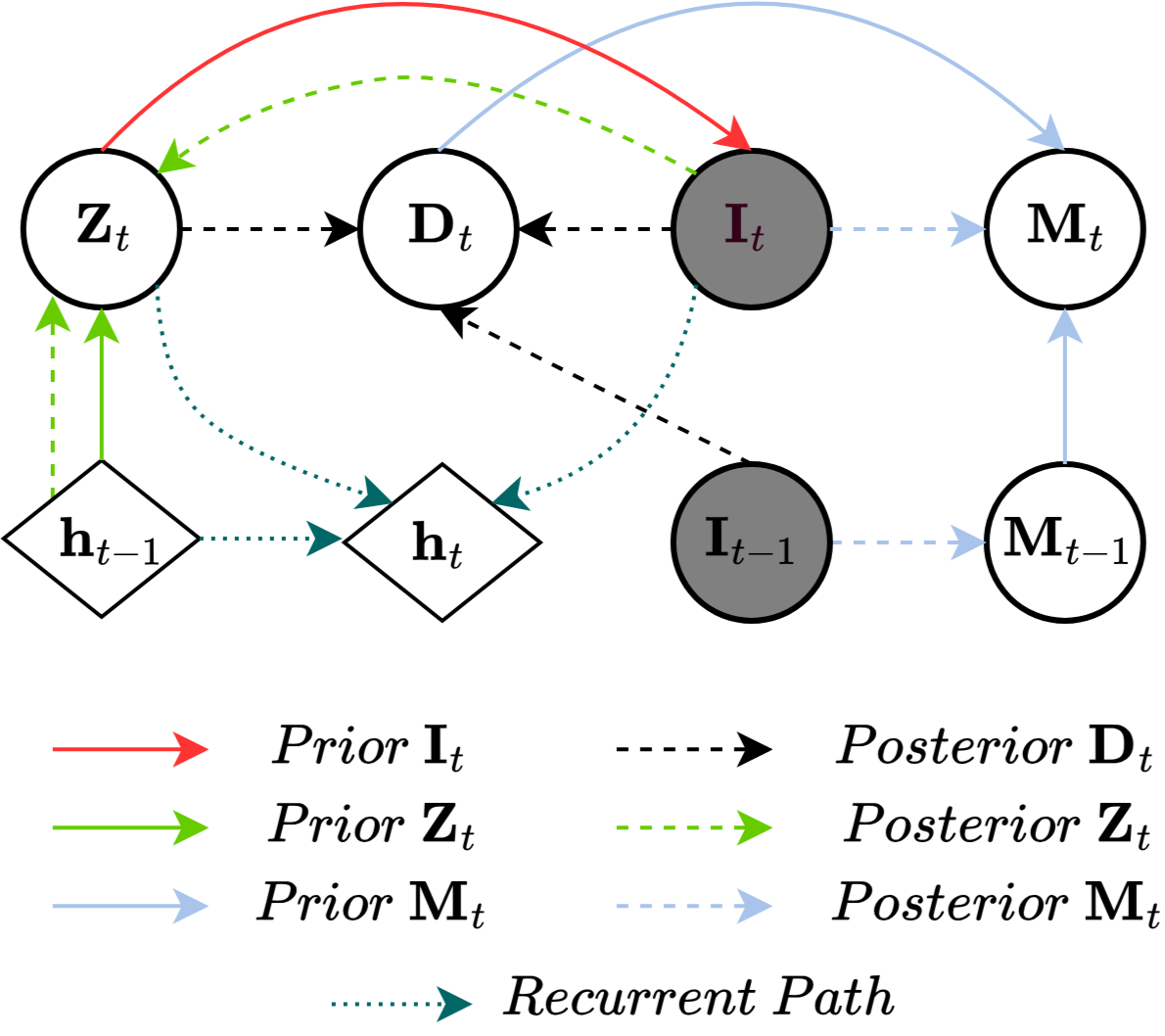}}
    \caption{Directed graphical model of the presented framework at time step $t$. Grey-filled nodes denote observed variables, while white nodes correspond to unobserved variables. Solid arrows indicate prior distributions: the latent variable $\mathbf{Z}_{t}$ is conditioned on the previous hidden state $\mathbf{h}_{t-1}$ (green), the SAX volume $\mathbf{I}_{t}$ is conditioned on $\mathbf{Z}_{t}$ (red), and the cardiac shape mask $\mathbf{M}_{t}$ is conditioned on $\mathbf{M}_{t-1}$ and DVF $\mathbf{D}_{t}$ (blue). Dashed arrows represent posterior distributions: $\mathbf{D}_{t}$ is inferred from $\mathbf{I}_{t}$, $\mathbf{I}_{t-1}$, and $\mathbf{Z}_{t}$; $\mathbf{Z}_{t}$ is inferred from $\mathbf{I}_{t}$ and $\mathbf{h}_{t-1}$; and $\mathbf{M}_{t}$ is inferred from $\mathbf{I}_{t}$. The dotted teal arrows indicate the recurrent path, where the hidden state $\mathbf{h}_{t}$ is updated using $\mathbf{h}_{t-1}$, $\mathbf{Z}_{t}$, and $\mathbf{I}_{t}$.}
  \label{fig1}
\end{figure}

Figure \ref{fig1} shows the directed graphical model of our Bayesian modelling with simplified joint and posterior distributions. As shown in Figure~\ref{fig1}, in the prior path, we condition $\mathbf{M}_{t}$ only on the previous shape mask $\mathbf{M}_{t-1}$ and the DVF at time $t$ $\mathbf{D}_{t}$. This reflects the fact that $\mathbf{M}_{t}$ can be estimated by warping $\mathbf{M}_{t-1}$ with $\mathbf{D}_{t}$, $\hat{\mathbf{M}}_{t} \approx \mathbf{M}_{t-1} \circ \mathbf{D}_{t}$, where $\hat{\mathbf{M}}_{t}$ is the warped shape mask and $\circ$ denotes the warping operator. We also condition the prior of $\textbf{I}_{t}$ on the latent variable $\mathbf{Z}_{t}$, assuming that this latent embedding encodes information about the SAX volumes across the cardiac sequence. To capture temporal dependencies across the SAX volume sequence, the proposed model employs a combination of 3D Convolutional layers and Long Short-Term Memory (LSTM), referred to as ConvLSTM. This network aggregates the dependencies among preceding SAX volumes $\textbf{I}_{<t}$ and latent variables $\textbf{Z}_{<t}$ through the hidden state variable, $\textbf{h}_{t-1} \in \mathbb{R}^{32 \times \frac{H}{16} \times \frac{W}{16} \times D}$, recursively. To perform this aggregation, the ConvLSTM model updates the hidden state at \textit{t} time, $\textbf{h}_{t}$, using the hidden state at the previous time, $\textbf{h}_{t-1}$, $\textbf{I}_{t}$, and $\textbf{Z}_{t}$, defined as $\textbf{h}_{t} = f_\phi(\textbf{h}_{t-1},\textbf{Z}_{t},\textbf{I}_{t})$, where $f_\phi$ is the ConvLSTM network with parameters $\phi$. The prior of $\textbf{Z}_{t}$ is conditioned on $\textbf{h}_{t-1}$ to enable the latent variable to learn the temporal evolution of the preceding SAX volumes and latent variables using the hidden state. For the prior of DVFs, we also assume a constant Gaussian distribution with zero mean and identity covariance, $p(\mathbf{D}_{t}) = \mathcal{N}(\mathbf{D}_{t}; \mathbf{0}, \mathbf{I})$ where $\mathcal{N}$, $\mathbf{0}$, and $\mathbf{I}$ denote the Gaussian distribution, zero, and identity matrices, respectively. This assumption ensures the computational tractability of the proposed model during training and optimisation. Based on the aforementioned assumptions, the joint probability is simplified as follows:
\begin{equation}\label{eq3}
    \mathcal{P} = \prod_{t=0}^{T-1} p(\textbf{M}_{t}| \textbf{M}_{t-1}, \textbf{D}_{t})
    p(\textbf{I}_{t}| \textbf{Z}_{t})
    p(\textbf{Z}_{t}| \textbf{h}_{t-1})
    p(\textbf{D}_{t}).
\end{equation}

For the simplification of the posterior distribution, $\mathcal{Q}$, as shown in Figure \ref{fig1}, the cardiac shape mask at each time point is conditioned on the SAX volume at the corresponding time, resulting in $q(\textbf{M}_{r}| \textbf{I}_{r})$ for the posterior distribution of cardiac shape masks, where  $0\leq r \leq T-1, r \notin \{t_{ED}, t_{ES}\}$. This assumption is made because cardiac masks can be directly estimated from the SAX volume using a segmentation network. The posterior distribution of the latent variable, $\textbf{Z}_t$, is also simplified by conditioning it on both the current SAX volume $\mathbf{I}_{t}$ and the hidden state $\mathbf{h}_{t-1}$. Conditioning on $\mathbf{I}_{t}$ enables $\mathbf{Z}_{t}$ to encode the spatial features of the SAX volume at each cardiac time point, whereas conditioning on $\mathbf{h}_{t-1}$ incorporates temporal information, which is accumulated from all preceding SAX volumes and latent embeddings through the ConvLSTM, into the latent representation. Consequently, $\mathbf{Z}_{t}$ can represent a latent embedding that reflects both spatial and temporal features. The posterior distribution of DVFs is also simplified by conditioning $\textbf{D}_{t}$ on the pair $(\mathbf{I}_{t}, \mathbf{I}_{t-1})$ together with $\mathbf{Z}_{t}$. This assumption enables the DVF posterior distribution to exploit features from sequential SAX volumes, as well as the learned spatio-temporal dependencies encoded in the latent space, to estimate DVFs for warping cardiac shape masks over the sequence. According to the above assumptions, the posterior distribution is simplified as follows:
\begin{equation}\label{eq4}
    \mathcal{Q} \!= \!\!\!\!\!\!\!\!\!\!\prod_{\substack{r=0 \\ r \notin \{t_{ED}, t_{ES}\}}}^{T-1} \!\!\!\!\!\!\!\!\!\!q(\textbf{M}_{r}| \textbf{I}_{r})
    \prod_{t=0}^{T-1} q(\textbf{D}_{t}| \textbf{I}_{t}, \textbf{I}_{t-1}, \textbf{Z}_{t})
    q(\textbf{Z}_{t}| \textbf{I}_{t}, \textbf{h}_{t-1}).
\end{equation}

Table \ref{tab:assumptions} provides the assumptions for the posterior and prior models. For the models listed in Table \ref{tab:assumptions}, $\boldsymbol{\mu}_{\textbf{D}_{t}} \!\!\in \!\!\mathbb{R}^{3 \times H \times W \times D}$ and  $\textbf{C}_{\textbf{D}_{t}}\!\! \in\!\! \mathbb{R}^{3 \times 3 \times H \times W \times D}$ denote the mean and covariance of $q(\textbf{D}_{t}| \textbf{I}_{t},\textbf{I}_{t-1},\textbf{Z}_{t})$, respectively. Similarly, $\boldsymbol{\mu}_{\textbf{Z}_{t,pi}}\!\! \in\!\! \mathbb{R}^{1 \times \frac{H}{16} \times \frac{W}{16} \times D}$, $\boldsymbol{\sigma}_{\textbf{Z}_{t,pi}}\!\! \in\!\! \mathbb{R}^{1 \times \frac{H}{16} \times \frac{W}{16} \times D}$, $\boldsymbol{\mu}_{\textbf{Z}_{t}}\!\! \in\!\! \mathbb{R}^{1 \times \frac{H}{16} \times \frac{W}{16} \times D}$, and $\boldsymbol{\sigma}_{\textbf{Z}_{t}}\!\! \in\!\! \mathbb{R}^{1 \times \frac{H}{16} \times \frac{W}{16} \times D}$ represent the means and standard deviations of $p(\textbf{Z}_{t}| \textbf{h}_{t-1})$ and $q(\textbf{Z}_{t}| \textbf{I}_{t},\textbf{h}_{t-1})$. It is assumed that $\mathbf{Z}_t$ is composed of $L$ independent latent vectors $\mathbf{z}_t^{(l)} ~l=1, 2, ..., L$, each following a Gaussian distribution with mean $\boldsymbol{\mu}_{\mathbf{z}_{t}}^{(l)}$ and variance $(\boldsymbol{\sigma}^2_{\mathbf{z}_{t}})^{(l)}$. For the prior and posterior distributions of anatomical masks, \textit{N} is the number of voxels and $c_{tnk}$ denotes the class label of \textit{k}\textsuperscript{th} class in \textit{n}\textsuperscript{th} voxel at \textit{t}\textsuperscript{th} time point.  $\boldsymbol{\boldsymbol{\theta}}_{tnk}$ and $\boldsymbol{\Psi}_{tnk}$ are the probabilities of \textit{k}\textsuperscript{th} class in \textit{n}\textsuperscript{th} voxel at \textit{t}\textsuperscript{th} time point obtained by $p(\textbf{M}_{t}| \textbf{M}_{t-1}, \textbf{D}_{t})$ and $q(\textbf{M}_{t}| \textbf{I}_{t})$, respectively. It is worth mentioning that all distributions except $p(\textbf{D}_{t})$, which is a fixed zero-mean Gaussian distribution, should be estimated in our modelling. We also assume $p(\textbf{I}_{t}|\textbf{Z}_{t}) = \mathcal{N}(\textbf{I}_{t}; ~\boldsymbol{\mu}_{I},\textbf{I})$ where $\boldsymbol{\mu}_{I}$ is the reconstructed SAX volume using $\textbf{Z}_{t}$, $\hat{\textbf{I}}_{t}(\textbf{Z}_{t})$.
\begin{table}[t!]
    \centering
    \fontsize{8pt}{10pt}\selectfont
    \caption{\fontsize{8pt}{10pt}\selectfont
    Assumptions of the proposed Bayesian model.
    This table summarises the prior and variational posterior distributions adopted for the latent variables, DVFs, cardiac shape masks, and SAX volumes within the probabilistic formulation.
    Gaussian distributions are used to model the latent variable $\mathbf{Z}_{t}$ and the DVF $\mathbf{D}_{t}$, while multinomial distributions are employed for modelling the cardiac shape masks $\mathbf{M}_{t}$.}
    \renewcommand{\arraystretch}{1.5}
    \begin{tabular}{c}
        \hline
        \hline
        $p(\textbf{D}_{t})=\mathcal{N}(\textbf{D}_{t}; \textbf{0}, \textbf{I})$ \\
        \hline
        $q(\textbf{D}_{t}| \textbf{I}_{t}, ~\textbf{I}_{t-1}, ~\textbf{Z}_{t}) = \mathcal{N}(\textbf{D}_{t}; \boldsymbol{\mu}_{\textbf{D}_{t}}, \textbf{C}_{\textbf{D}_{t}})$ \\
        \hline
        $p(\textbf{Z}_{t}| \textbf{h}_{t-1}) = \mathcal{N}(\textbf{Z}_{t}; ~\boldsymbol{\mu}_{\textbf{Z}_{t,pi}}, ~\boldsymbol{\sigma}^{2}_{{\textbf{Z}_{t,pi}}})=$ \\
        $\prod_{l=1}^{L} \mathcal{N}(\textbf{z}^{(l)}_{t}; ~\boldsymbol{\mu}^{(l)}_{\textbf{z}_{t,pi}}, ~(\boldsymbol{\sigma}^{2}_{{\textbf{z}_{t,pi}}})^{(l)})$ \\
        \hline
        $q(\textbf{Z}_{t}| \textbf{I}_{t}, ~\textbf{h}_{t-1}) = \mathcal{N}(\textbf{Z}_{t}; ~\boldsymbol{\mu}_{\textbf{Z}_{t}}, ~\boldsymbol{\sigma}^{2}_{{\textbf{Z}_{t}}})=$ \\
        $\prod_{l=1}^{L} \mathcal{N}(\textbf{z}^{(l)}_{t}; ~\boldsymbol{\mu}^{(l)}_{\textbf{z}_{t}}, ~(\boldsymbol{\sigma}^{2}_{{\textbf{z}_{t}}})^{(l)})$ \\
        \hline
        $p(\textbf{M}_{t}| \textbf{M}_{t-1}, \textbf{D}_{t}) = \prod_{n=1}^{N}\prod_{k=1}^{K}(\hat{\textbf{M}_{t}})_{nk}^{c_{tnk}}= \prod_{n=1}^{N}\prod_{k=1}^{K}\boldsymbol{\theta}_{tnk}^{c_{tnk}}$ \\
        \hline
        $q(\textbf{M}_{t}| \textbf{I}_{t}) =\prod_{n=1}^{N}\prod_{k=1}^{K}\boldsymbol{\Psi}_{tnk}^{c_{tnk}}$ \\
        \hline
        $p(\textbf{I}_{t}|\textbf{Z}_{t}) = \mathcal{N}(\textbf{I}_{t}; ~\boldsymbol{\mu}_{I},\textbf{I})$\\
        \hline
        \hline
    \end{tabular}
    \label{tab:assumptions}
\end{table}

\subsection{Framework's Architecture}
\begin{figure}[b!]
    \centering
    \includegraphics[width=0.8\textwidth]{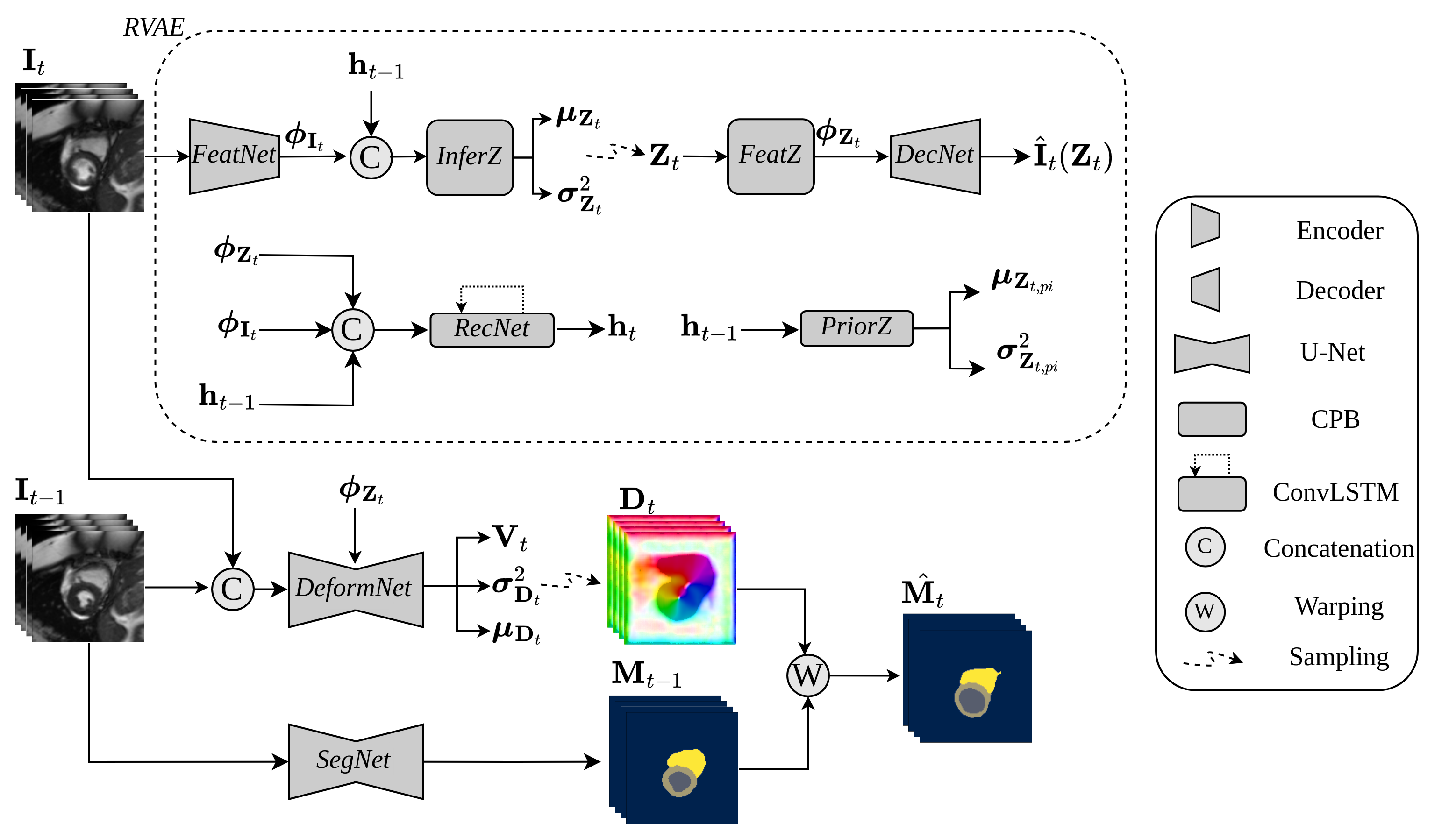}
    \caption{\fontsize{8pt}{10pt}\selectfont
    Inference architecture of \textit{CardioMorphNet} framework for cardiac shape registration.
    Channel Processing Blocks (CPBs) extract hierarchical features while preserving the spatial resolution of feature maps, whereas encoder--decoder modules perform spatial downsampling and upsampling along the height and width dimensions. (\textit{RVAE}) models spatio-temporal dependencies by inferring the latent variable $\mathbf{Z}_{t}$ from the current SAX volume $\mathbf{I}_{t}$ and the previous hidden state $\mathbf{h}_{t-1}$, and updates the recurrent state $\mathbf{h}_{t}$ accordingly.
    Here, $\boldsymbol{\phi}_{\mathbf{I}_{t}}$ denotes the feature map extracted from $\mathbf{I}_{t}$ by \textit{FeatNet}, while $\boldsymbol{\phi}_{\mathbf{Z}_{t}}$ denotes the feature representation derived from the latent variable $\mathbf{Z}_{t}$ via \textit{FeatZ}.
    \textit{DeformNet} estimates the DVF $\mathbf{D}_{t}$ using two consecutive SAX volumes $\mathbf{I}_{t}$ and $\mathbf{I}_{t-1}$ together with the latent variable $\mathbf{Z}_{t}$.
    The \textit{SegNet} module outputs the cardiac shape mask $\mathbf{M}_{t-1}$, which is subsequently warped using the estimated DVF $\mathbf{D}_{t}$ to obtain the registered shape $\hat{\mathbf{M}}_{t}$.}
    \label{fig_blockdigram}
\end{figure}

Figure \ref{fig_blockdigram} illustrates the main block diagram of the proposed Bayesian cardiac shape registration framework. As depicted in Figure \ref{fig_blockdigram}, the proposed framework comprises a \textit{RVAE} model for modelling spatio-temporal dependencies, a segmentation branch (\textit{SegNet}) for cardiac regions segmentation, and a registration module (\textit{DeformNet}) for estimating DVFs. More details on the deep learning architecture of the proposed framework are provided in \ref{app:architecture}.

\textit{RVAE}'s encoder extracts spatial features from each SAX volume at time \textit{t}, $\boldsymbol{\phi}_{\mathbf{I}_{t}}\!\in \!\mathbb{R}^{32 \times \frac{H}{16} \times \frac{W}{16} \times D}$, using \textit{FeatNet}, and \textit{InferZ} combines these spatial features with the hidden state, $\mathbf{h}_{t-1}$, providing the temporal information to estimate the mean and variance of the posterior distribution $q(\textbf{Z}_{t}| \textbf{I}_{t},\textbf{h}_{t-1})$. Given $\boldsymbol{\mu}_{\textbf{Z}_{t}}$ and $\boldsymbol{\sigma}_{\textbf{Z}_{t}}$, the latent embedding, $\textbf{Z}_{t}$, is sampled using the reparameterisation trick \cite{kingma2013auto}. Since the latent embedding is estimated based on $\boldsymbol{\phi}_{\mathbf{I}_{t}}$ and $\mathbf{h}_{t-1}$, it captures both spatial and temporal features throughout the cardiac cycle. \textit{PriorZ} also estimates the parameters of the prior distribution $p(\mathbf{Z}_{t}|\mathbf{h}_{t-1})$ using the hidden state variable and allows the latent embedding to learn the temporal evolution of the sequence. The decoder reconstructs $\hat{\mathbf{I}}_{t}(\mathbf{Z}_{t})$ through \textit{FeatZ} and \textit{DecNet}. \textit{RecNet}, which is a ConvLSTM-based network, serves as the memory component of the \textit{RVAE}, updating the hidden state recursively via $\mathbf{h}_{t-1}$,  $\boldsymbol{\phi}_{\mathbf{I}_{t}}$, and the features of the latent variable at time \textit{t}, $\boldsymbol{\phi}_{\mathbf{Z}_{t}}\! \in \!\mathbb{R}^{32 \times \frac{H}{16} \times \frac{W}{16} \times D}$, to capture temporal dependencies and integrate them into the hidden state variable.

\textit{DeformNet} is a 3D U-Net network modelling the posterior distribution of the DVF at time point \textit{t}, $\textbf{D}_{t}$. The U-Net structure of \textit{DeformNet} enables learning multi-scale deformation features from the concatenation of $\textbf{I}_{t}$ and $\textbf{I}_{t-1}$ for estimating $\textbf{D}_{t}$. It also employs spatio-temporal dependencies for DVF estimation by concatenating the features of the latent embedding, $\boldsymbol{\phi}_{\mathbf{Z}_{t}}$, with embedded deformation features. \textit{DeformNet} outputs $\boldsymbol{\mu}_{\textbf{D}_{t}}$, $\textbf{V}_{t}\! \in\! \mathbb{R}^{3 \times H \times W \times D}$, and $\boldsymbol{\sigma}^{2}_{{\textbf{D}_{t}}}\!\! \in \mathbb{R}^{1 \times H \times W \times D}$. $\mathbf{V}_{t}$ is a low-rank tensor capturing correlations between displacement components, and $\boldsymbol{\sigma}^{2}_{\mathbf{D}_{t}}$ represents the voxel-wise variance of the DVF at time $t$. Using $\mathbf{V}_{t}$ and $\boldsymbol{\sigma}^{2}_{\mathbf{D}_{t}}$, the covariance of the DVF posterior at time $t$ is estimated as $\mathbf{C}_{\mathbf{D}_{t}} = \boldsymbol{\sigma}^{2}_{\mathbf{D}_{t}}\mathbf{I} + \mathbf{V}_{t}^{\intercal}\mathbf{V}_{t}$, where $(\cdot)^{\intercal}$ denotes the transpose operator. Next, $\textbf{D}_{t}$ is estimated using the reparameterisation trick, given $\boldsymbol{\mu}_{\textbf{D}_{t}}$ and $\mathbf{C}_{\mathbf{D}_{t}}$. The explicit parametric modelling of DVF covariances derived from Bayesian modelling provides direct, voxel-resolved access to DVF uncertainties.

\textit{SegNet} is a 3D U-Net network that estimates cardiac shape masks as segmentation probability maps. It models the posterior distribution of the segmentation mask at time $t$, $\mathbf{M}_{t}$, given the corresponding SAX volume $\mathbf{I}_{t}$, i.e., $q(\mathbf{M}_{t}| \mathbf{I}_{t})$. The prior over the cardiac masks, $p(\mathbf{M}_{t}|\mathbf{M}_{t-1}, \mathbf{D}_{t})$, is defined through a warping operator that registers the mask at the previous time step, $\mathbf{M}_{t-1}$, to time $t$ using the estimated DVF $\mathbf{D}_{t}$, such that $\hat{\mathbf{M}}_{t} \approx \mathbf{M}_{t-1} \circ \mathbf{D}_{t}$, where $\hat{\mathbf{M}}_{t}$ denotes the warped mask and $\circ$ is the warping operator. Both $q(\mathbf{M}_{t}|\mathbf{I}_{t})$ and $p(\mathbf{M}_{t} | \mathbf{M}_{t-1}, \mathbf{D}_{t})$ are modelled as voxel-wise multinomial distributions with time-resolved parameters $\Psi_{tnk}$ and $\theta_{tnk}$, respectively (see Table~\ref{tab:assumptions}). Crucially, due to our Bayesian formulation, DVF estimation is supervised by matching the current shape mask $\mathbf{M}_{t}$, predicted by \textit{SegNet}, to the warped previous mask $\hat{\mathbf{M}}_{t}$ via minimising $D_{KL}(q(\mathbf{M}_{t}| \mathbf{I}_{t})\parallel p(\mathbf{M}_{t}|\mathbf{M}_{t-1}, \mathbf{D}_{t}))$, where $D_{KL}$ denotes the Kullback--Leibler divergence (KL divergence)---see $\mathcal{L}_{\text{semi-shape}}$ in Eq.~\eqref{eq8}. Minimising this term encourages the DVFs to warp $\mathbf{M}_{t-1}$ so that $\hat{\mathbf{M}}_{t}$ aligns with \textit{SegNet}'s predicted shape, thereby guiding motion estimation to focus on the cardiac anatomical regions.

\subsection{Objective Function}
To estimate $\mathcal{Q}$ and $\mathcal{P}$, the negative of the variational Evidence Lower Bound (ELBO) \cite{kingma2013auto} loss called $\mathcal{L}_{ELBO}$ is minimised. Bending energy loss, $\mathcal{L}_{Smooth}$, is also needed to control the smoothness of DVFs \cite{yu2016back}. Therefore, the overall loss function is given by:
\begin{equation}\label{eq5}
    \mathcal{L}= \mathcal{L}_{ELBO} + \rho\mathcal{L}_{Smooth},
\end{equation}
where $\mathcal{L}$ and $\rho$ are the total loss and the bending energy regularisation coefficient, respectively. In this study, $\rho$ is set to 0.03. $\mathcal{L}_{Smooth}$ is computed using $\sum_{t=1}^{T}\lVert \nabla \textbf{D}_{t}\rVert^2$, where $\lVert \cdot \rVert^2$ and $\nabla$ are the squared $\ell_2$-norm and the spatial gradient operator, respectively. $\mathcal{L}_{ELBO}$ is defined as $\mathbb{E}_{\mathcal{Q}}\left[\log\left(\frac{\mathcal{Q}}{\mathcal{P}}\right)\right]$, where $\mathbb{E}_{\mathcal{Q}}$ is the expectation operator with respect to $\mathcal{Q}$. Considering the assumed factorisation of $\mathcal{Q}$ and $\mathcal{P}$, this results in a form given by:
\begin{equation}\label{eq6}
    \mathcal{L}_{ELBO} = \mathcal{L}_{shape} +\lambda_1\mathcal{L}_{D} + \lambda_2\mathcal{L}_{Z}+ \lambda_3\mathcal{L}_{rec},
\end{equation}
where $\mathcal{L}_{shape}$ is the shape cross-entropy loss, $\mathcal{L}_{rec}$ is the reconstruction loss between $\hat{\textbf{I}}_{t}(\textbf{Z}_{t})$ and $\textbf{I}_{t}$, $\mathcal{L}_{D}$ and $\mathcal{L}_{Z}$ are the KL divergence terms $D_{KL}(q(\textbf{D}_{t}| \textbf{I}_{t}, \textbf{I}_{t-1}, \textbf{Z}_{t})\parallel p(\mathbf{D}_{t}))$ and $D_{KL}(q(\textbf{Z}_{t}| \textbf{I}_{t}, \textbf{h}_{t-1})\parallel p(\mathbf{Z}_{t}|\textbf{h}_{t-1}))$, respectively. $\lambda_i, i=1,2,3$ in Eq. \eqref{eq6} are the regularisation coefficients for balancing the values of loss terms. To train the model, $\lambda_1$, $\lambda_2$, and $\lambda_3$ are set to $10^{-4}$, $2\times10^{-4}$, and $0.3$, respectively, and are chosen empirically to achieve optimal performance.

$\mathcal{L}_{shape}= \mathcal{L}_{sup-shape} + \mathcal{L}_{semi-shape}$, where $\mathcal{L}_{sup-shape}$ is the supervised cross-entropy between the ground-truth masks at $t_{\mathrm{ED}}$ and $t_{\mathrm{ES}}$ time steps and the warped segmentation maps at these frames, and $\mathcal{L}_{semi-shape}$ is the semi-supervised cross-entropy between the segmentation maps and the warped ones at time points other than $t_{\mathrm{ED}}$ and $t_{\mathrm{ES}}$. $\mathcal{L}_{semi-shape}$ and $\mathcal{L}_{sup-shape}$ are derived from $D_{KL}(q(\mathbf{M}_{t}| \mathbf{I}_{t})\parallel p(\mathbf{M}_{t}|\mathbf{M}_{t-1}, \mathbf{D}_{t}))$ and $\mathbb{E}_{\mathcal{Q}}\left[\sum_{r} \log\left(p(\textbf{M}_{r} | \textbf{M}_{r-1}, \textbf{D}_{r})\right)\right]$ where $r \in \{t_{ED}, t_{ES}\}$, respectively. These terms are given by
\begin{equation}\label{eq7}
    \mathcal{L}_{sup-shape} = -\!\!\!\!\!\!\!\!\!\sum_{r=\{t_{ED},t_{ES}\}}\sum_{n=1}^{N} \sum_{k=1}^{K} c_{rnk}\log\left(\boldsymbol{\theta}_{rnk}\right),
\end{equation}
\begin{equation}\label{eq8}
    \mathcal{L}_{semi-shape} = \!\!\!\!\!\!\!\!\!\sum_{\substack{t=1 \\ t \notin \{t_{ED}, t_{ES}\}}}^{T}\!\!\!\!\!\!\!\!\!\sum_{k=1}^{K}\sum_{n=1}^{N}\boldsymbol{\Psi}_{tnk}\left(\log(\boldsymbol{\Psi}_{tnk})-\log(\boldsymbol{\theta}_{tnk})\right),
\end{equation}
where $N$ is the number of voxels, $K$ is the number of classes in the cardiac masks, $\boldsymbol{\Psi}_{tnk}$ is the probability of class $k$ at voxel $n$ and time $t$ predicted by $q(\mathbf{M}_{t}|\mathbf{I}_{t})$, and $\boldsymbol{\theta}_{tnk}$ is the corresponding probability given by $p(\mathbf{M}_{t}|\mathbf{M}_{t-1}, \mathbf{D}_{t})$. Both $\mathcal{L}_{sup-shape}$ and $\mathcal{L}_{semi-shape}$ enable our framework to supervise DVFs by matching the warped masks, which are obtained by $p(\textbf{M}_{t}| \textbf{M}_{t-1}, \textbf{D}_{t})$, with the ground-truth masks at $t_{\mathrm{ED}}$ and $t_{\mathrm{ES}}$ time points and with the estimated masks by $q(\textbf{M}_{t}|\textbf{I}_{t})$ at other time steps.

$\mathcal{L}_{D}$ and $\mathcal{L}_{Z}$ are also given by:
\begin{equation}\label{eq9}
    \mathcal{L}_{D} = \frac{1}{2}\sum_{t=1}^{T} \left( \log\left(\left|\textbf{C}_{{\textbf{D}_{t}}}\right|\right) + 3
            + \boldsymbol{\mu}_{\textbf{D}_{t}}^\intercal \boldsymbol{\mu}_{\textbf{D}_{t}}
            + \operatorname{Tr}\left\{\textbf{C}_{\textbf{D}_{t}}\right\} \right),
\end{equation}
\begin{equation}\label{eq10}
    \mathcal{L}_{Z} = \frac{1}{2}\sum_{t=1}^{T}\sum_{i=1}^{L}(\log(\boldsymbol{\sigma}^{2}_{\textbf{z}_{t,pi}})^{(i)}-\log(\boldsymbol{\sigma}^{2}_{\textbf{z}_{t}})^{(i)}- 1 + \frac{(\boldsymbol{\sigma}^{2}_{\textbf{z}_{t}})^{(i)} + (\boldsymbol{\mu}^{(i)}_{\textbf{z}_{t}} - \boldsymbol{\mu}^{(i)}_{\textbf{z}_{t,pi}})^2}{(\boldsymbol{\sigma}^{2}_{\textbf{z}_{t,pi}})^{(i)}}),
\end{equation}
where $|\cdot|$ and $\operatorname{Tr}\{\cdot\}$ denote the determinant and the matrix trace, respectively. $\mathcal{L}_{rec}$ is also derived as:
\begin{equation}\label{eq11}
    \mathcal{L}_{rec} =\frac{1}{L_{Z}}\sum_{t=1}^{T}\sum_{l=1}^{L_{Z}}\left\lVert \textbf{I}_{t}- \hat{\textbf{I}}_{t}(\textbf{Z}^{(l)}_{t})\right\rVert^2,
\end{equation}
where $L_{Z}$ is the number of Monte Carlo samples of $\textbf{Z}_{t}$ derived from $q(\textbf{Z}_{t}| \textbf{I}_{t},\textbf{h}_{t-1})$ and $\hat{\textbf{I}}_{t}(\textbf{Z}^{(l)}_{t})$ is the reconstructed SAX volume using \textit{l}\textsuperscript{th} sample of $\textbf{Z}_{t}$ at time step \textit{t}.

\subsection{Training and Inferencing Pipeline}
The proposed framework is trained using a two-stage procedure. In the first stage, the \textit{SegNet} module is trained and validated on a subset of our dataset to achieve the desired cardiac anatomical regions segmentation performance, while the other modules are frozen. The cross-entropy between the ground-truth masks and the outputs of \textit{SegNet} is used as the loss function to train this module. Next, the layers of \textit{SegNet} are frozen, and the weights of the remaining modules are unfrozen and trained using the overall loss function provided in Eq. \eqref{eq5} for learning cardiac motion. This two-phase training procedure enables the model to extract multi-scale deformations from sequential SAX volumes and to supervise the estimated DVFs to concentrate on the cardiac anatomical regions with the aid of the pre-trained \textit{SegNet} module.

During inference, two sequential SAX volumes are fed into the framework throughout the cardiac cycle. First, $\boldsymbol{\phi}_{\textbf{Z}_{t}}$ is computed given the SAX volume at time \textit{t}, $\textbf{I}_{t}$, and $\textbf{h}_{t-1}$. Then, the cardiac shape mask at time \textit{t-1}, $\textbf{M}_{t-1}$, is estimated using \textit{SegNet}. Next, given two sequential SAX volumes, $\textbf{I}_{t}$ and $\textbf{I}_{t-1}$, and $\boldsymbol{\phi}_{\textbf{Z}_{t}}$, the DVF at time \textit{t}, $\textbf{D}_{t}$, is estimated and $\textbf{M}_{t-1}$ is warped using $\textbf{D}_{t}$ to obtain the warped shape mask at time \textit{t}, $\hat{\textbf{M}_{t}}$. This procedure runs at each cardiac time step to obtain the estimated cardiac motions throughout the cardiac cycle.

\section{Experiments and Results}\label{sec4}
\subsection{Experimental Setup}
This study uses SAX data from 4,000 randomly selected subjects in the UK Biobank dataset. A total of 1,000 subjects are used to develop the \textit{SegNet}, and the remaining subjects are used to develop the full framework. For both the \textit{SegNet} module and the full framework, each data subset is split into 80\% training, 10\% validation, and 10\% testing subsets. Each SAX volume contains 50 frames per subject. The ED and ES time steps and ground-truth masks at these time points are obtained using an automated tool provided in \cite{bai2018automated}. This tool has been used in several studies to obtain ground-truth masks at the mentioned cardiac time points \cite{meng2023deepmesh, meng2022mesh, meng2022mulvimotion}. We further evaluate the proposed framework on the M\&M dataset \cite{campello2021multi}. To do this, the trained model is fine-tuned on the training and validation sets and evaluated on the test subset of the M\&M dataset.

For preprocessing, the SAX sequences are downsampled to six frames to reduce computational load while maintaining sufficient cardiac motion dynamics. The downsampled SAX sequence begins at the ED time point, includes two successive SAX volumes leading to the ES time point, and is followed by two additional frames to complete the cardiac cycle. Slices of each SAX volume are then cropped around the heart using the ground-truth mask at the ED frame with a 30-pixel margin, resized to 128$\times$128, and normalised to the range [0,1]. The normalised slices are stacked along the depth dimension, and the results are zero-padded to 16 slices for uniform depth.

The whole framework is implemented using Python, PyTorch, and MONAI \cite{cardoso2022monai}\footnote{Code available at: \url{https://github.com/rezamovahed93/CardioMorphNet.git}}. The Adam optimiser with a learning rate of $10^{-3}$ is used to train the model in both training phases. The experiments were conducted on a machine with 8 CPU cores, 32 GB of memory and an NVIDIA A5000 GPU. The batch size for both training phases is set to 8.

\subsection{Evaluation and Assessment}
To evaluate the performance of the proposed framework for cardiac motion estimation using cine SAX data, several conventional and state-of-the-art learning-based image registration methods are implemented for comparison. Specifically, free-form registration techniques, including LCC-Demons~\cite{lorenzi2013lcc} and Symmetric Normalisation (SyN) implemented in Advanced Normalisation Tools (ANTs)~\cite{avants2011reproducible}, as well as state-of-the-art learning-based methods, such as VoxelMorph~\cite{balakrishnan2019voxelmorph}, TLRN~\cite{wu2024tlrn}, TCSF~\cite{wei2020temporal}, DragNet~\cite{zakeri2023dragnet}, JMS~\cite{qin2018joint}, and SegMorph~\cite{bi2024segmorph}, are implemented and evaluated on the same datasets. To assess cardiac motion estimation accuracy, the Dice Similarity Coefficient (DSC), Jaccard Index (JAC), mean surface distance (MSD), and 95th-tile Hausdorff Distance (HD95) are computed using ground-truth masks in ED and ES frames and the corresponding sequentially warped cardiac shapes. The diffeomorphic property is measured by the number of voxels with the negative Jacobian determinant (NJD) \cite{ashburner2007fast}. Motion uncertainty assessment is conducted for our method and other probabilistic approaches to evaluate their confidence in DVF estimation within the cardiac region. To show the benefits of the improved motion prediction for cardiac image registration techniques, the clinical indices such as the volume of the LV and RV at ED and ES (LVEDV, LVESV, RVEDV and RVESV), LV volume over the cardiac cycle (LVV), the stroke volume of LV and RV (LVSV, RVSV), the left ventricular muscle mass (LVMM), and the ejection fraction of the LV and RV (LVEF and RVEF) are computed using the cardiac shapes obtained by the methods, and the root mean squared error (RMSE) is measured between these variables and their reference values computed using the ground-truth masks at ED and ES frames. An ablation study is also performed to demonstrate the effect of the components of the proposed framework for cardiac motion estimation. It is also analysed how the selection of the bending energy regularisation coefficient affects both cardiac shape registration and diffeomorphic property. The paired t-test is used in the evaluations to assess whether differences between methods are statistically significant.

\subsection{Cardiac Shape Registration Assessment}

\begin{figure}[b!]
    \centering
    \includegraphics[width=0.80\textwidth]{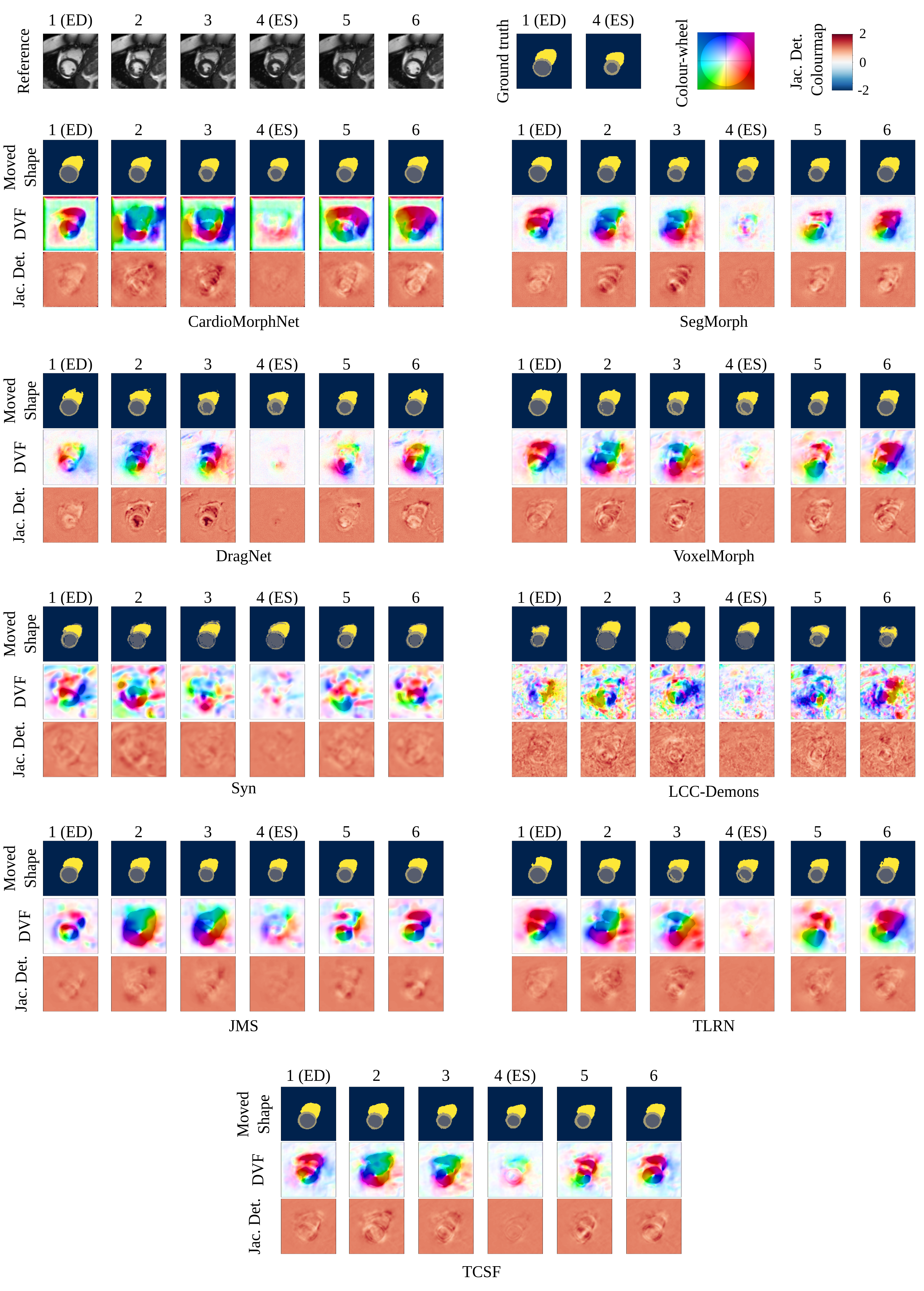}
    \caption{Qualitative comparison of cardiac shape registration results obtained using the proposed method and other approaches. Results are shown across multiple cardiac phases from ED to ES and vice versa. For each method, the warped cardiac shapes (Moved Shape), the DVFs, and the corresponding Jac. Det. maps are visualised. DVFs are shown using a colour-wheel representation, while Jac. Det. maps are illustrated using a colour map ranging from $-2$ to $2$. As illustrated, the proposed method outputs cardiac shapes that are better aligned with cardiac anatomical regions and yields DVFs that represent more physiologically plausible cardiac motion than those of other methods. It also demonstrates that the DVFs obtained by our method are sufficiently smooth.}
    \label{fig_re1}
\end{figure}

\begin{figure}[t!]
    \centering
    \includegraphics[width=0.65\textwidth]{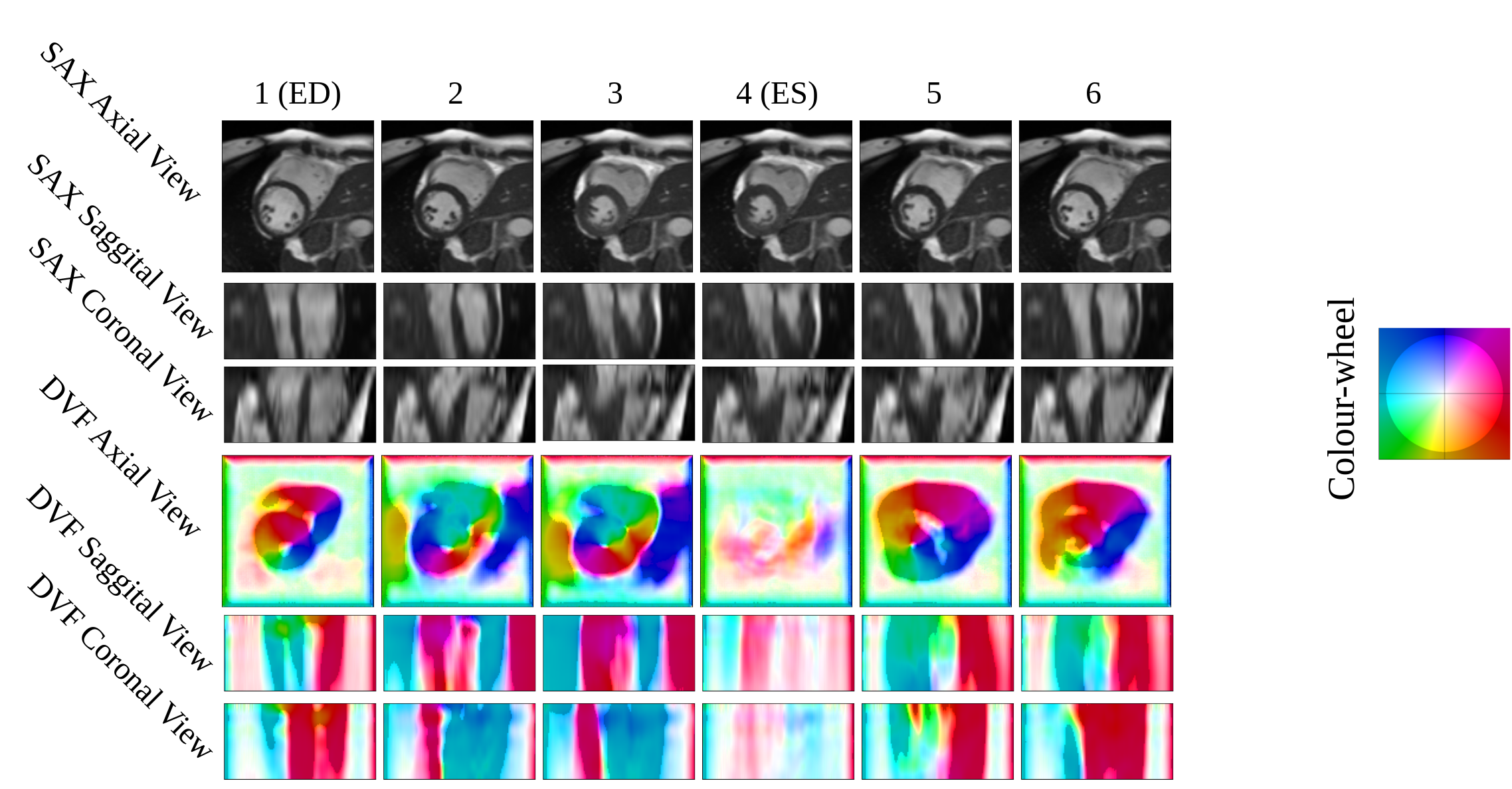}
    \caption{Visualisation of the DVFs estimated by CardioMorphNet in axial, sagittal, and coronal views. DVFs are shown using a colour-wheel representation based on the $x$--$y$, $y$--$z$, and $x$--$z$ displacement components for the axial, sagittal, and coronal planes, respectively.
    The results demonstrate that the proposed method estimates ventricular motion consistently across all three orthogonal views, indicating its ability to capture coherent three-dimensional cardiac deformation throughout the entire volume.}
    \label{fig_re_dvf}
\end{figure}

Figure \ref{fig_re1} shows a representative example of the obtained cardiac shape masks, DVFs, and Jacobian determinants (Jac. Dets.) by CardioMorphNet and other methods throughout the cardiac cycle. As shown, the moved shape masks obtained by CardioMorphNet, JMS, TCSF, and SegMorph are more aligned with the LV, RV, and LVmyo regions and exhibit higher similarity to the ground-truth masks at ED and ES frames than those achieved by baseline approaches. These observations indicate that methods equipped with a segmentation module perform better in cardiac shape registration than others. It is also shown in Figure \ref{fig_re1} that DragNet, VoxelMorph, and TLRN perform comparably to each other in terms of cardiac shape registration based on the similarity between the moved shapes in ED and ES phases and the ground-truth masks at these time points, but are less accurate than our framework, JMS, TCSF, and SegMorph. The remaining methods provide the poorest cardiac shape registration performance because of the lowest similarity between the moved shapes obtained by these methods and the ground-truth masks. The DVFs in Figure \ref{fig_re1} demonstrate that SyN and LCC-Demons provide the least precise cardiac motion estimation compared with others, since their DVFs contain various motion fields not related to the cardiac region and its boundaries. DragNet, VoxelMorph, and TLRN offer more accurate DVF estimation than the aforementioned methods. Our framework, JMS, SegMorph, and TCSF achieve the most precise DVF estimation for cardiac motion, particularly our method, which accurately estimates heart rotation at the ED, fifth, and sixth time points. It is also observed that our method estimated motions around the cardiac region at the second and third time points, whereas other approaches did not. This is due to the nonlinear heart motion from ED to ES, which is driven by myocardial contraction and torsion. Figure \ref{fig_re_dvf} also shows a sample of the 3D DVFs estimated by our method in axial, sagittal, and coronal views over the cardiac cycle. As shown, our method estimates ventricular motion consistently in the coronal, sagittal, and axial planes, indicating its ability to capture coherent three-dimensional cardiac deformation across the entire volume. The visualised Jac. Dets. in Figure \ref{fig_re1} indicate that our method, SegMorph, TCSF, DragNet, and VoxelMorph can better focus on cardiac anatomical regions for DVF estimation while estimating smooth DVFs.

\begin{figure}[t!]
    \centering
    \includegraphics[width=0.7\textwidth]{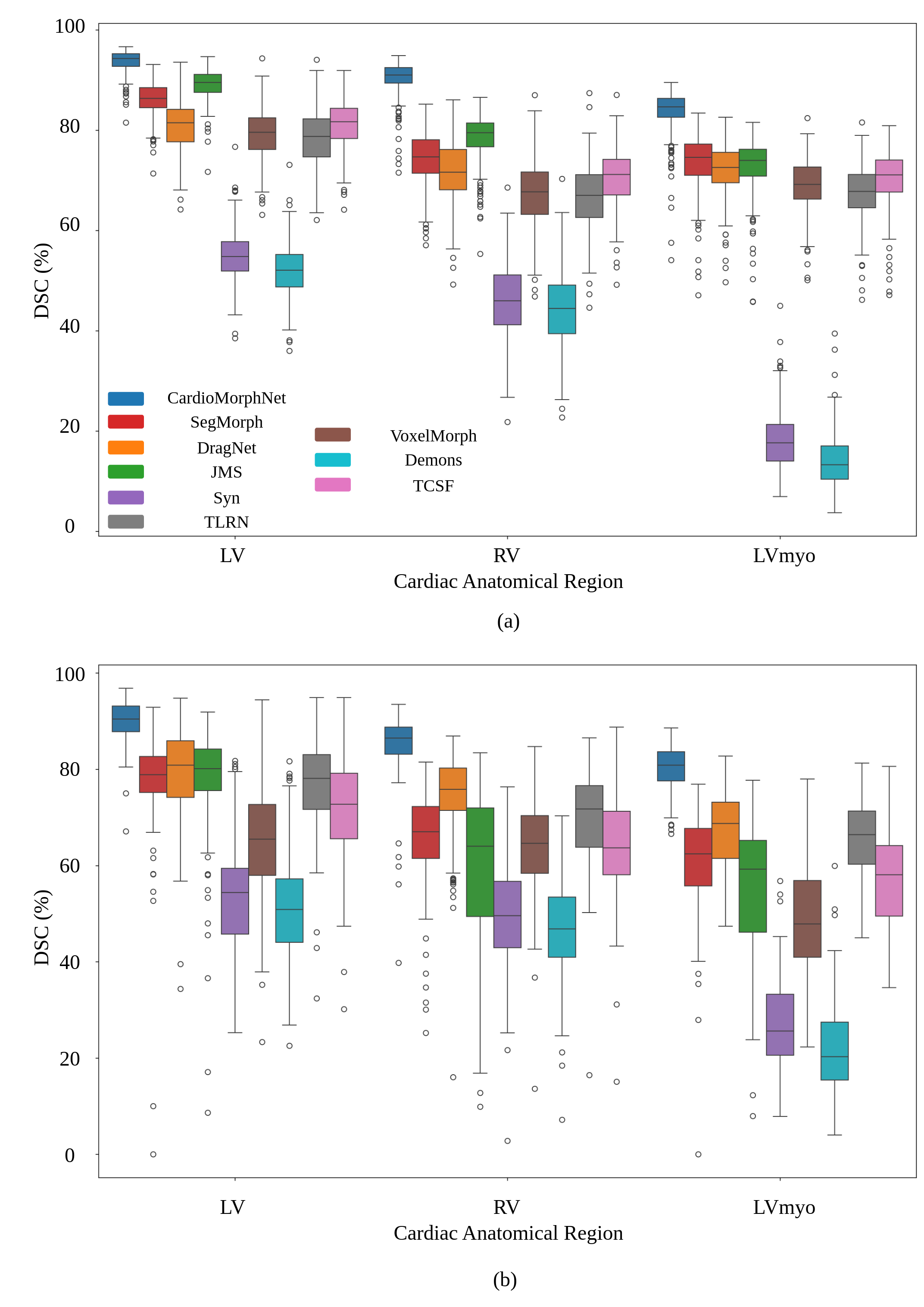}
    \caption{Boxplot distributions of the percentage of (\%) DSC values for cardiac shape registration obtained using the proposed framework and other methods across different cardiac anatomical regions. Panel (a) presents the results on the UK Biobank dataset, while panel (b) shows the corresponding results on the M\&M dataset.
    As illustrated, the proposed method consistently achieves higher median DSC values and lower performance variability across all anatomical regions and datasets than the other approaches, indicating more robust and accurate cardiac shape registration.}
    \label{fig_boxplot}
\end{figure}

Tables \ref{table.3} and \ref{table.3_MM} report the quantitative results of cardiac shape registration using the UK Biobank and M\&M datasets, respectively. According to Tables \ref{table.3} and \ref{table.3_MM}, the presented method achieves higher mean DSC and JAC and lower mean HD95 and MSD metrics than other state-of-the-art methods, indicating that our framework provides the best performance for cardiac shape registration. For the UK Biobank dataset, SegMorph, JMS, and TCSF achieve results most comparable to those of the proposed framework. In the M\&M dataset, DragNet, SegMorph, JMS, and TLRN achieve performance comparable to that of the presented method for cardiac shape registration. Other methods (e.g., LCC-Demons and SyN) yield the lowest DSC and JAC means, and the highest HD95 and MSD means, suggesting they provide the poorest cardiac shape registration performance. The boxplots of the DSC metric percentage obtained by our framework and other methods for both datasets are shown in Figure \ref{fig_boxplot}. As demonstrated, CardioMorphNet achieves the highest median DSC with the narrowest interquartile range, indicating that our method outperforms others for cardiac shape registration in terms of accuracy and robustness. It also shows that the DSC values obtained by all methods for LVmyo shape registration are lower than those for other cardiac regions. This suggests that LVmyo motion estimation is more challenging than that of other cardiac regions, as its anatomical structure is thinner and more complex. With respect to NJD, DragNet obtains the smoothest DVFs among all methods by achieving the lowest NJD mean across both datasets, whereas our framework achieves slightly higher NJD means compared with DragNet. This reflects a trade-off between accurate boundary alignment of discrete cardiac structures and strict diffeomorphic regularisation. Since our framework is explicitly supervised using cardiac shape masks, it prioritises anatomically consistent shape mask alignment, which may locally relax smoothness constraints compared with strongly regularised intensity-based approaches. Similar observations have been reported in previous studies \cite{chen2022joint, pace2013locally, hua2017multiresolution}. Importantly, despite this trade-off, our method achieves improved anatomical alignment, as demonstrated by the DSC, JAC, and HD95 metrics in Tables \ref{table.3} and \ref{table.3_MM}, and the NJD values remain within a comparable range to those of existing learning-based registration methods. Overall, the obtained results for cardiac shape mask registration throughout the cardiac cycle demonstrate that CardioMorphNet is more accurate and robust than others for heart motion estimation.

\begin{table}[t!]
  \caption{Quantitative comparison of cardiac shape registration results on the UK Biobank dataset in terms of the percentage (\%) of DSC and JAC metrics and HD95 and MSD values in millimetres (mm), computed for ED and ES time points between cardiac shape ground-truth masks and the moved ones by methods. NJD is also reported for assessing the diffeomorphic property. All metrics are reported as Mean$\pm$Std. \textbf{Bold} items indicate the best results.}\label{table.3}
  \centering
  \resizebox{0.8\textwidth}{!}{%
    \begin{tabular}{c c c c c c c}
      \hline
      \hline
      Method & Region of Interest & DSC(\%) & JAC(\%) & HD95 (mm) & MSD (mm) & \#NJD \\
      \hline
        \multirow{3}{*}{CardioMorphNet}
        & LV     & \textbf{93.77$\pm$2.11} & \textbf{88.35$\pm$3.63} & \textbf{1.44$\pm$0.65} & \textbf{0.29$\pm$0.15} & \multirow{3}{*}{566.85$\pm$310.25} \\
        & LVmyo  & \textbf{83.84$\pm$4.20} & \textbf{72.39$\pm$5.69} & \textbf{1.50$\pm$0.49} & \textbf{0.33$\pm$0.13} & {} \\
        & RV     & \textbf{90.42$\pm$3.20} & \textbf{82.65$\pm$5.00} & \textbf{1.50$\pm$0.88} & \textbf{0.33$\pm$0.18} & {} \\
      \hline
        \multirow{3}{*}{VoxelMorph}
        & LV     & 79.20$\pm$4.91 & 65.83$\pm$6.67 & 3.76$\pm$1.50 & 1.03$\pm$0.34 & \multirow{3}{*}{837.31$\pm$470.33} \\
        & LVmyo  & 68.98$\pm$5.27 & 52.90$\pm$6.04 & 2.61$\pm$0.52 & 0.66$\pm$0.14 & {} \\
        & RV     & 67.20$\pm$6.55 & 50.97$\pm$7.37 & 7.19$\pm$3.24 & 1.58$\pm$0.61 & {} \\
      \hline
        \multirow{3}{*}{SegMorph}
        & LV     & 86.16$\pm$3.28 & 75.82$\pm$4.96 & 2.90$\pm$1.49 & 0.76$\pm$0.34 & \multirow{3}{*}{580.45$\pm$263.07} \\
        & LVmyo  & 73.77$\pm$5.32 & 58.71$\pm$6.38 & 2.32$\pm$0.61 & 0.57$\pm$0.15 & {} \\
        & RV     & 74.28$\pm$5.31 & 59.36$\pm$6.60 & 5.15$\pm$2.59 & 1.14$\pm$0.46 & {} \\
      \hline
        \multirow{3}{*}{DragNet}
        & LV     & 80.67$\pm$4.91 & 67.88$\pm$6.77 & 3.77$\pm$1.59 & 0.97$\pm$0.38 & \multirow{3}{*}{\textbf{244.60$\pm$256.76}} \\
        & LVmyo  & 72.02$\pm$4.81 & 56.49$\pm$5.68 & 2.48$\pm$0.54 & 0.60$\pm$0.13 & {} \\
        & RV     & 71.51$\pm$6.06 & 55.99$\pm$7.25 & 5.76$\pm$3.05 & 1.26$\pm$0.53 & {} \\
      \hline
        \multirow{3}{*}{LCC-Demons}
        & LV     & 51.93$\pm$4.95 & 35.22$\pm$4.56 & 18.05$\pm$5.21 & 4.46$\pm$1.60 & \multirow{3}{*}{--} \\
        & LVmyo  & 13.92$\pm$5.30 & 7.57$\pm$3.17 & 30.35$\pm$13.35 & 7.64$\pm$3.22 & {} \\
        & RV     & 44.30$\pm$7.25 & 28.73$\pm$6.03 & 12.74$\pm$5.75 & 3.43$\pm$1.89 & {} \\
      \hline
        \multirow{3}{*}{SyN}
        & LV     & 54.68$\pm$4.82 & 37.78$\pm$4.63 & 14.45$\pm$6.05 & 3.86$\pm$1.38 & \multirow{3}{*}{--} \\
        & LVmyo  & 18.09$\pm$5.76 & 10.06$\pm$3.59 & 31.59$\pm$12.39 & 7.68$\pm$3.83 & {} \\
        & RV     & 45.72$\pm$7.29 & 29.92$\pm$6.13 & 12.36$\pm$5.09 & 3.24$\pm$1.63 & {} \\
      \hline
        \multirow{3}{*}{JMS}
        & LV     & 85.53$\pm$3.31 & 74.86$\pm$4.91 & 2.50$\pm$1.22 & 0.69$\pm$0.24 & \multirow{3}{*}{702.06$\pm$435.78} \\
        & LVmyo  & 64.32$\pm$4.92 & 47.60$\pm$5.15 & 2.94$\pm$0.60 & 0.78$\pm$0.15 & {} \\
        & RV     & 72.96$\pm$4.79 & 57.64$\pm$5.74 & 4.91$\pm$3.20 & 1.12$\pm$0.69 & {} \\
      \hline
        \multirow{3}{*}{TLRN}
        & LV     & 78.47$\pm$5.44 & 64.90$\pm$7.34 & 3.63$\pm$1.30 & 1.04$\pm$0.31 & \multirow{3}{*}{1015.15$\pm$503.95} \\
        & LVmyo  & 67.48$\pm$5.53 & 51.18$\pm$6.19 & 2.62$\pm$0.51 & 0.69$\pm$0.14 & {} \\
        & RV     & 66.71$\pm$6.66 & 50.42$\pm$7.47 & 7.08$\pm$3.09 & 1.61$\pm$0.59 & {} \\
      \hline
        \multirow{3}{*}{TCSF}
        & LV     & 82.18$\pm$3.72 & 69.92$\pm$5.32 & 2.75$\pm$1.05 & 0.84$\pm$0.24 & \multirow{3}{*}{668.34$\pm$386.52} \\
        & LVmyo  & 68.00$\pm$3.93 & 51.65$\pm$4.53 & 2.67$\pm$0.49 & 0.70$\pm$0.12 & {} \\
        & RV     & 69.75$\pm$4.52 & 53.73$\pm$5.28 & 4.96$\pm$3.04 & 1.24$\pm$0.64 & {} \\
      \hline
      \hline
    \end{tabular}
  }
\end{table}

\begin{table}[t!]
\caption{Quantitative comparison of cardiac shape registration results on the M\&M dataset in terms of the percentage (\%) of DSC and JAC metrics and HD95 and MSD values in millimetres (mm), computed for ED and ES time points between cardiac shape ground-truth masks and the moved ones by methods. NJD is also reported for assessing the diffeomorphic property. All metrics are reported as Mean$\pm$Std. \textbf{Bold} items indicate the best results.}
\label{table.3_MM}
\centering
\resizebox{0.8\textwidth}{!}{%
\begin{tabular}{c c c c c c c}
\hline
\hline
Method & Region of Interest & DSC(\%) & JAC(\%) & HD95 (mm) & MSD (mm) & \#NJD \\
\hline
\multirow{3}{*}{CardioMorphNet}
& LV    & \textbf{89.99$\pm$4.18} & \textbf{82.05$\pm$6.57} & \textbf{2.22$\pm$2.08} & \textbf{0.55$\pm$0.37} & \multirow{3}{*}{772.79$\pm$396.36} \\
& LVmyo & \textbf{80.36$\pm$4.56} & \textbf{67.40$\pm$6.23} & \textbf{1.98$\pm$0.75} & \textbf{0.49$\pm$0.18} & {} \\
& RV    & \textbf{84.94$\pm$6.92} & \textbf{74.36$\pm$8.96} & \textbf{2.53$\pm$2.84} & \textbf{0.64$\pm$0.70} & {} \\
\hline
\multirow{3}{*}{VoxelMorph}
& LV    & 65.54$\pm$12.64 & 50.08$\pm$14.41 & 9.05$\pm$4.67 & 2.48$\pm$1.36 & \multirow{3}{*}{143.08$\pm$204.11} \\
& LVmyo & 49.30$\pm$11.26 & 33.48$\pm$10.25 & 5.39$\pm$1.63 & 1.49$\pm$0.48 & {} \\
& RV    & 64.01$\pm$10.09 & 47.84$\pm$10.51 & 9.52$\pm$6.28 & 2.19$\pm$1.93 & {} \\
\hline
\multirow{3}{*}{SegMorph}
& LV    & 77.40$\pm$11.17 & 64.15$\pm$11.57 & 5.28$\pm$4.82 & 1.41$\pm$1.29 & \multirow{3}{*}{689.15$\pm$417.88} \\
& LVmyo & 60.80$\pm$10.09 & 44.36$\pm$9.42 & 5.39$\pm$6.12 & 1.43$\pm$2.48 & {} \\
& RV    & 65.29$\pm$9.95 & 49.21$\pm$10.08 & 9.87$\pm$9.69 & 2.32$\pm$2.20 & {} \\
\hline
\multirow{3}{*}{DragNet}
& LV    & 79.55$\pm$9.07 & 66.91$\pm$11.59 & 4.10$\pm$2.58 & 1.16$\pm$0.68 & \multirow{3}{*}{\textbf{0.53$\pm$1.38}} \\
& LVmyo & 67.31$\pm$7.77 & 51.23$\pm$8.73 & 3.15$\pm$0.89 & 0.84$\pm$0.23 & {} \\
& RV    & 74.00$\pm$9.30 & 59.49$\pm$10.42 & 4.94$\pm$5.36 & 1.28$\pm$1.91 & {} \\
\hline
\multirow{3}{*}{LCC-Demons}
& LV    & 50.93$\pm$11.42 & 34.98$\pm$10.74 & 29.48$\pm$16.54 & 6.56$\pm$3.93 & \multirow{3}{*}{--} \\
& LVmyo & 22.13$\pm$9.49 & 12.79$\pm$6.46 & 29.74$\pm$14.37 & 6.82$\pm$3.13 & {} \\
& RV    & 46.77$\pm$10.40 & 31.12$\pm$8.82 & 13.76$\pm$9.62 & 3.90$\pm$4.56 & {} \\
\hline
\multirow{3}{*}{SyN}
& LV    & 53.74$\pm$11.37 & 37.60$\pm$11.11 & 30.66$\pm$17.45 & 6.56$\pm$3.99 & \multirow{3}{*}{--} \\
& LVmyo & 27.32$\pm$9.36 & 16.17$\pm$6.57 & 30.31$\pm$14.36 & 6.53$\pm$3.31 & {} \\
& RV    & 49.28$\pm$10.69 & 33.36$\pm$9.40 & 12.77$\pm$9.42 & 3.56$\pm$4.54 & {} \\
\hline
\multirow{3}{*}{JMS}
& LV    & 77.45$\pm$11.97 & 64.45$\pm$13.02 & 9.87$\pm$11.76 & 2.14$\pm$2.22 & \multirow{3}{*}{179.69$\pm$463.40} \\
& LVmyo & 55.78$\pm$13.47 & 39.82$\pm$12.30 & 9.27$\pm$7.27 & 2.39$\pm$2.04 & {} \\
& RV    & 59.54$\pm$16.69 & 44.26$\pm$15.86 & 15.01$\pm$13.91 & 3.56$\pm$3.53 & {} \\
\hline
\multirow{3}{*}{TLRN}
& LV    & 77.15$\pm$9.70 & 63.75$\pm$12.22 & 4.97$\pm$2.95 & 1.33$\pm$0.72 & \multirow{3}{*}{381.62$\pm$728.44} \\
& LVmyo & 65.47$\pm$8.17 & 49.21$\pm$8.98 & 3.40$\pm$0.99 & 0.90$\pm$0.26 & {} \\
& RV    & 69.99$\pm$9.68 & 54.63$\pm$10.76 & 7.03$\pm$5.99 & 1.63$\pm$1.88 & {} \\
\hline
\multirow{3}{*}{TCSF}
& LV    & 71.88$\pm$10.80 & 57.19$\pm$13.01 & 6.35$\pm$3.29 & 1.76$\pm$0.87 & \multirow{3}{*}{155.19$\pm$493.63} \\
& LVmyo & 56.93$\pm$9.86 & 40.46$\pm$9.73 & 4.03$\pm$1.14 & 1.12$\pm$0.31 & {} \\
& RV    & 63.98$\pm$10.38 & 47.87$\pm$11.02 & 8.27$\pm$5.97 & 2.02$\pm$2.00 & {} \\
\hline
\hline
\end{tabular}
}
\end{table}

\subsection{Uncertainty Assessment}
\begin{figure}[t!]
    \centering
    \includegraphics[width=0.5\textwidth]{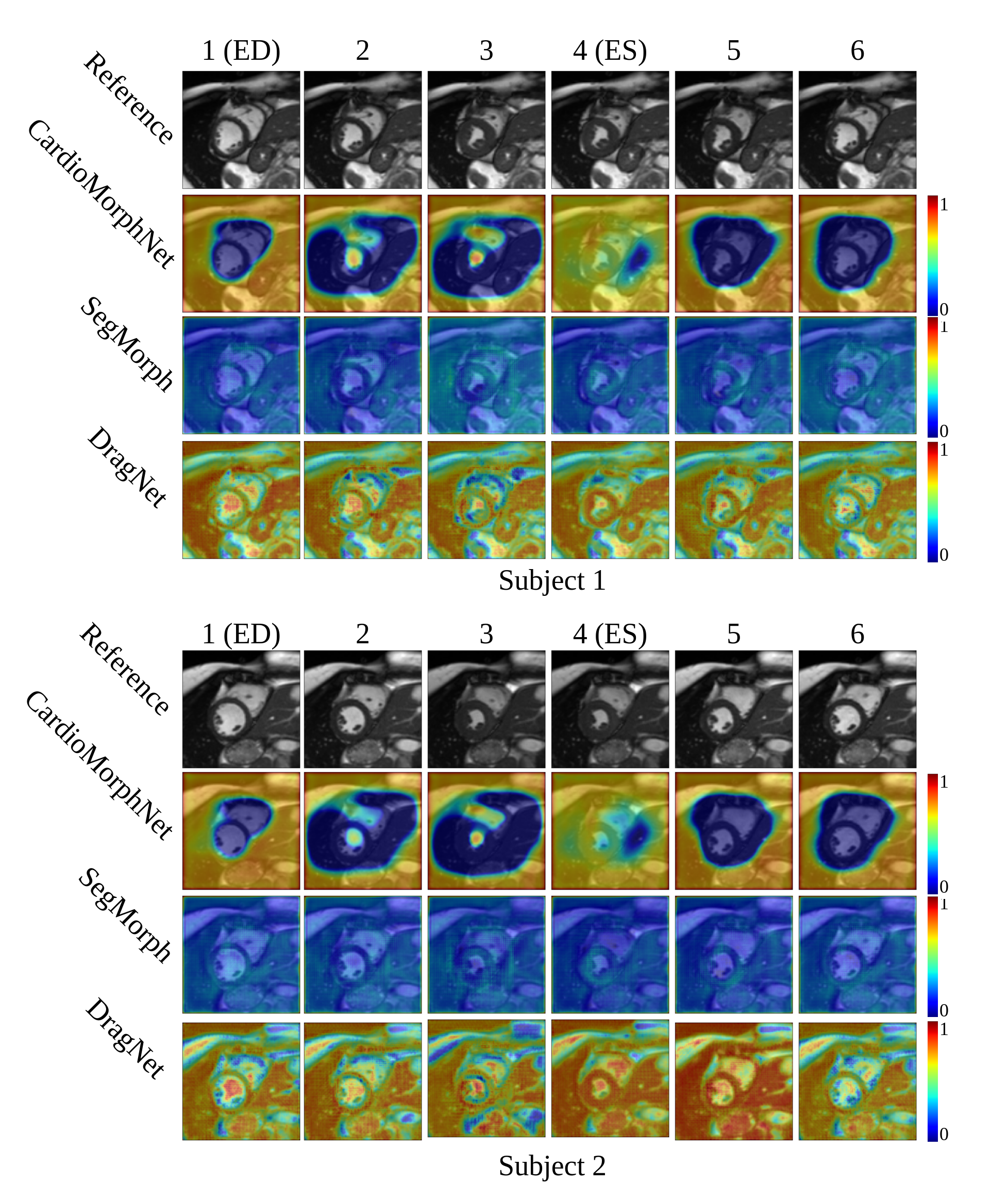}
    \caption{Visualisation of DVF uncertainty maps for the middle SAX slice of two representative subjects obtained using the proposed framework, SegMorph, and DragNet. Each uncertainty map is overlaid on the corresponding original SAX view, with values normalised to the range $[0, 1]$.
    As illustrated, the proposed framework yields DVFs with lower uncertainty values that are more spatially concentrated within the cardiac regions compared to the other probabilistic registration methods.}
    \label{fig_uncmaps}
\end{figure}

Figure \ref{fig_uncmaps} illustrates the overlaid DVF uncertainty maps on the corresponding SAX slices for two subjects, obtained using the proposed framework, SegMorph and DragNet, which are based on probabilistic modelling. DVF uncertainty is quantified using  $H(\textbf{D}_{t})\approx0.5\log((2\pi)^{3}|\textbf{C}_{\textbf{D}_{t}}|)$, where $H(\mathbf{D}_t)$ denotes the differential entropy of the Gaussian posterior $q(\mathbf{D}_t|\cdot)$, and $\mathbf{C}_{\mathbf{D}_t}$ is the covariance matrix of DVFs. Lower uncertainty values indicate greater confidence in motion estimation. As shown in Figure \ref{fig_uncmaps}, the low uncertainty values of the DVFs obtained by CardioMorphNet are more concentrated on the cardiac region compared to other probabilistic registration methods, at the ED frame and at the fifth and sixth time points, suggesting that the proposed framework focuses more precisely on the cardiac region than SegMorph and DragNet.

It is observed that the uncertainty values of our framework in the intermediate systolic frames (second and third) are higher in the LV and RV regions than in the surrounding regions. This is because cardiac motion across the systolic transition between the ED and ES is most rapid and nonlinear, driven by myocardial contraction, thickening, and torsion \cite{young1994three, hansen1988torsional}. Unlike ED and ES, for which ground-truth segmentation masks are available and strongly constrain the DVFs, these intermediate frames are supervised only by the semi-supervised shape consistency term. As a result, the posterior distribution of the DVFs naturally exhibits higher uncertainty within the LV and RV regions during these frames, where motion is both physiologically complex and less directly constrained.

Notably, the lower uncertainty observed outside the cardiac region in these frames is a desirable property of the proposed framework. Since CardioMorphNet supervises DVFs based on anatomical shape consistency rather than intensity-based similarity, background regions are encouraged to remain static, resulting in low posterior variance outside the heart. This contrasts with intensity-driven methods such as DragNet and SegMorph, which tend to distribute uncertainty more uniformly across the image. The fact that uncertainty decreases again in later frames demonstrates temporal stability and consistency of the learned motion model, rather than divergence. Overall, this behaviour indicates that our method appropriately localises uncertainty to anatomically meaningful regions during phases of complex motion, which is a key advantage of the proposed shape-guided Bayesian formulation.

\begin{figure}[t!]
    \centering
    \includegraphics[width=0.6\textwidth]{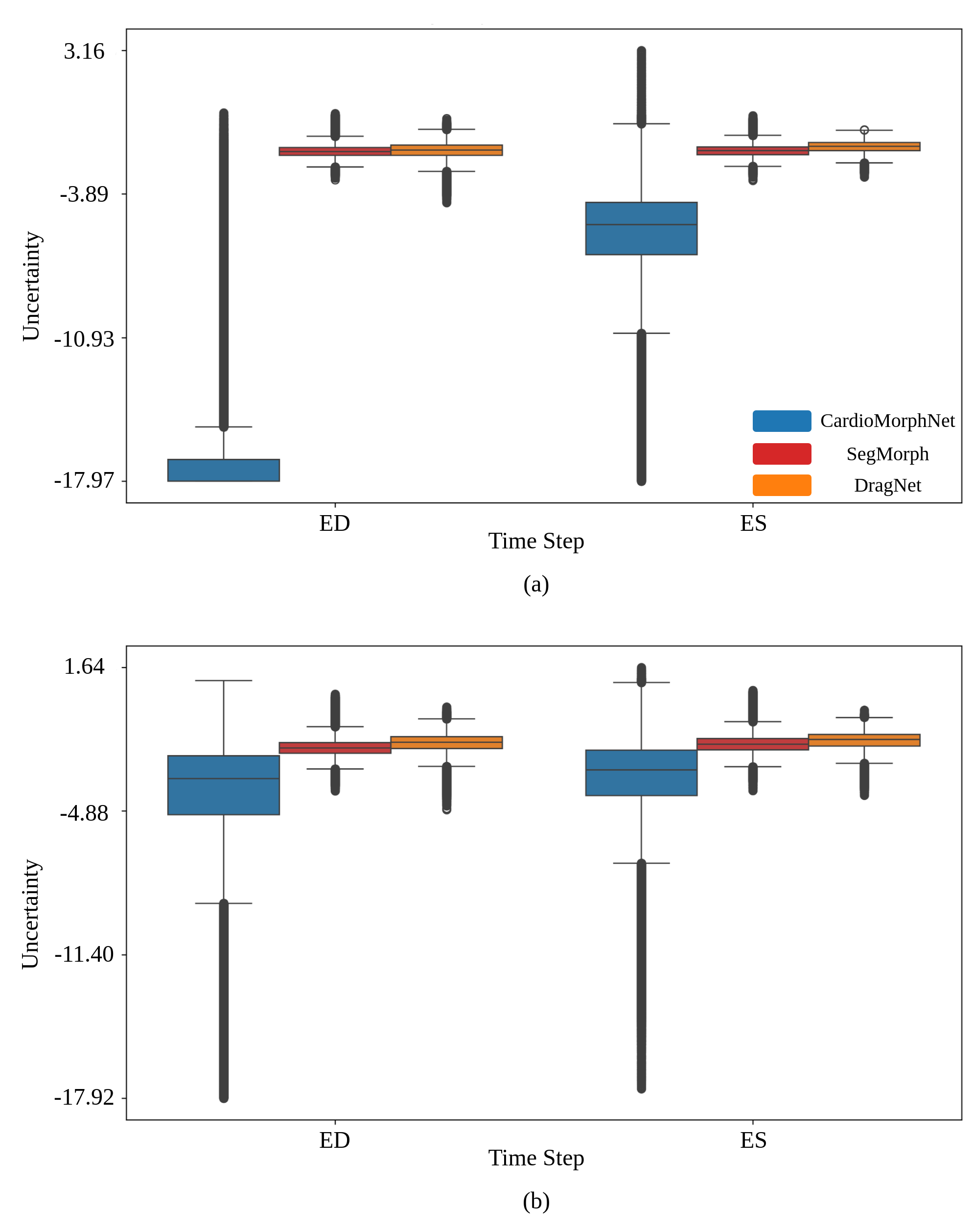}
    \caption{Boxplot distributions of uncertainty values within the cardiac region at ED and ES obtained using the proposed framework, SegMorph, and DragNet.
    Panel (a) presents the results on the UK Biobank dataset, while panel (b) shows the corresponding results on the M\&M dataset.
    As illustrated, the proposed framework consistently yields lower uncertainty levels and reduced variability at both ED and ES time points compared to the competing methods, indicating more confident and stable DVFs estimation within the cardiac region.}
    \label{fig_boxplot_unc}
\end{figure}

The boxplots of the DVF uncertainty values within the cardiac region at the ED and ES time points, obtained using our approach and other methods, are shown in Figure \ref{fig_boxplot_unc}. As illustrated in Figure \ref{fig_boxplot_unc}, the DVF uncertainty values obtained by the presented framework for the UK Biobank dataset are significantly lower than those obtained by SegMorph and DragNet for both time steps (p-value $<$ 0.001). For the M\&M dataset, the distribution of uncertainty values from our method is closer to those of DragNet and SegMorph, but remains lower than theirs. Moreover, Figure \ref{fig_boxplot_unc} shows that the DVF uncertainty values obtained by our framework from the UK Biobank dataset at the ES time point are higher than those at the ED time point, indicating that motion estimation in the ES phase is more challenging for our method than in the ED phase. These results indicate that our method is more confident and robust in DVF estimation in the cardiac region, and it achieves superior performance in focusing on the cardiac anatomical structure and its boundaries during DVF estimation compared with SegMorph and DragNet.

\subsection{Evaluation of Clinical Indices Extraction}
Tables \ref{table.ci_ukb} and \ref{table.ci_mm} report the RMSEs of estimated clinical indices obtained by the proposed framework and other methods using the UK Biobank and M\&M datasets, respectively. Overall, CardioMorphNet achieves the lowest or among the lowest RMSE values across almost all LV and RV volumetric indices, including LVEDV, LVESV, RVEDV, RVESV, LVSV, RVSV, myocardial mass (LVMM), and functional indices (LVEF, RVEF) on both datasets, indicating superior agreement with the reference values. On the UK Biobank dataset, CardioMorphNet outperforms other approaches across most clinical indices, with substantial improvements in ventricular volumes and ejection fractions, while TCSF and VoxelMorph provide more accurate estimates of LVESV and LVMM.

\begin{figure}[t!]
    \centering
    \includegraphics[width=1.0\textwidth]{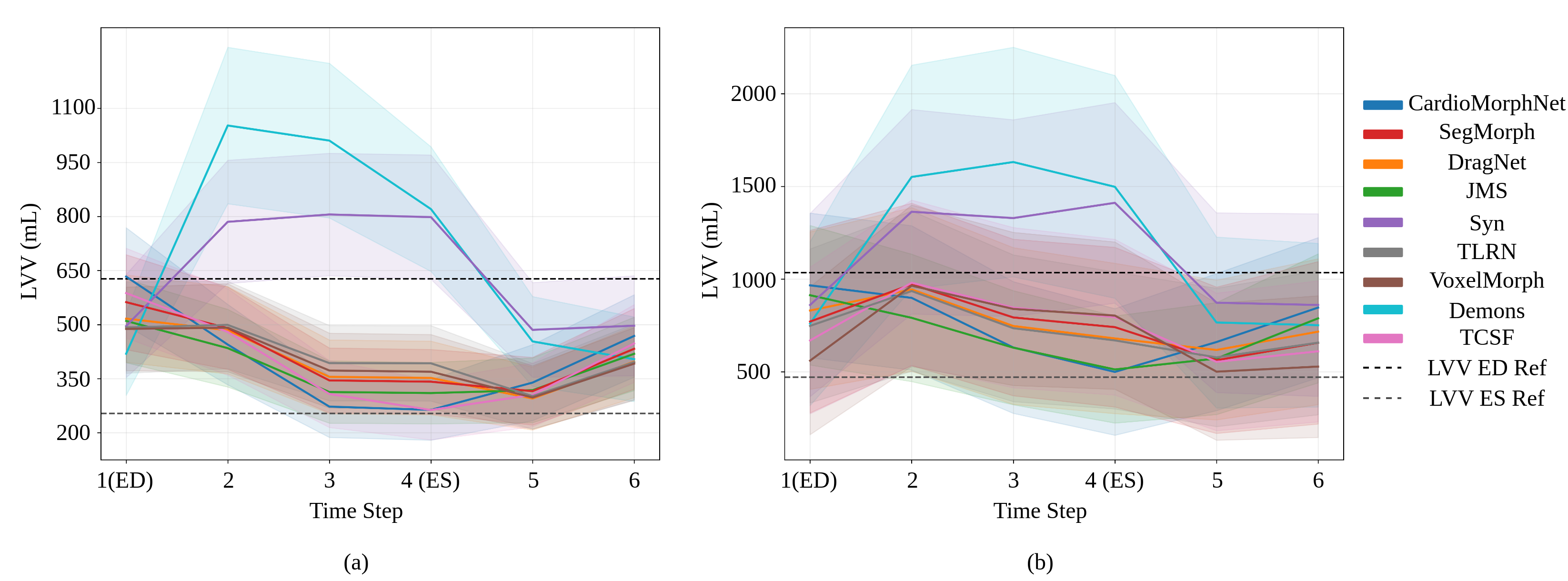}
    \caption{\fontsize{8pt}{10pt}\selectfont
    LVV trajectories estimated over the cardiac cycle for the test subsets of (a) the UK Biobank dataset and (b) the M\&M dataset using the cardiac shape masks obtained by the proposed framework and competing methods.
    Solid lines represent the mean LVV values across all test subjects, while the shaded regions indicate the corresponding standard deviation, reflecting inter-subject variability.
    The two dashed horizontal lines denote the reference LVV values at ED and ES, computed from the ground-truth shape masks of the test subsets.
    As illustrated, the proposed framework outputs LVV curves that more closely follow the reference values at both ED and ES and exhibit smoother temporal evolution across intermediate frames compared to the other methods, indicating more accurate and temporally consistent cardiac motion estimation.}
    \label{fig_LVV}
\end{figure}

Figure \ref{fig_LVV} illustrates the LVV parameter values over the cardiac cycle, estimated by the presented framework and other approaches, on both datasets. As shown in Figure \ref{fig_LVV}, our framework closely follows the physiological LVV pattern, capturing the expected decrease from ED to ES and subsequent recovery, while maintaining a slight deviation from the reference lines and comparatively lower variance across time. In contrast, other methods exhibit larger fluctuations, over- or under-estimation at intermediate frames. Their LVV values are farther from the reference values at the ED and ES time points than those obtained by our method. Notably, traditional registration-based approaches such as SyN and LCC-Demons exhibit inconsistencies in LVV estimation across the cardiac cycle, whereas learning-based methods show greater temporal consistency.

It is worth mentioning that the clinical indices used in this study are static parameters that can be directly obtained using segmentation networks. To further investigate, a comparison between the proposed motion estimation framework and the segmentation baselines U-Net \cite{ronneberger2015u} and nnU-Net \cite{isensee2021nnu} regarding estimating these indices is provided in \ref{app:seg_results}. As the evaluation using motion-related biomarkers, such as myocardial strain quantification, is not feasible in this study due to the lack of tagged MRI data and reliable ground-truth strain measurements, these clinical indices are used since their reference values can be computed from the ground-truth masks provided in the datasets. Although these indices do not directly measure motion quality, they provide useful evidence that the estimated motion preserves anatomically meaningful cardiac shapes and yields clinically consistent functional measurements. Therefore, these results indicate that the proposed cardiac motion estimation framework achieves more robust and accurate performance in estimating clinical cardiac parameters than other motion estimation approaches.

\subsection{Ablation Study}
To conduct an ablation study of the proposed framework, the \textit{SegNet} and \textit{RVAE} components are selectively enabled or disabled across four configurations, and their performance is evaluated on the cardiac shape registration task throughout the cardiac cycle. In all these configurations, \textit{DeformNet} is enabled to perform DVF estimation. Tables \ref{table.4} and \ref{table.4mm} list the DSC and HD95 metrics for cardiac shape registration obtained by all configurations on the UK Biobank and M\&M datasets, respectively. For configuration 2, where \textit{RVAE} is enabled and \textit{SegNet} is disabled, we design the prior as  $\mathcal{P} = \prod_{t=0}^{T-1}p(\textbf{I}_{t}| \textbf{I}_{t-1}, \textbf{D}_{t})p(\textbf{Z}_{t}| \textbf{h}_{t-1})p(\textbf{D}_{t})$, and the posterior as $\mathcal{Q} = \prod_{t=0}^{T-1} q(\textbf{D}_{t}| \textbf{I}_{t}, \textbf{I}_{t-1}, \textbf{Z}_{t})q(\textbf{Z}_{t}| \textbf{I}_{t}, \textbf{h}_{t-1})$, while the prior and posterior for configuration 4 are designed as $\mathcal{P} = \prod_{t=0}^{T-1}p(\textbf{I}_{t}| \textbf{I}_{t-1}, \textbf{D}_{t})p(\textbf{D}_{t})$ and $\mathcal{Q} = \prod_{t=0}^{T-1} q(\textbf{D}_{t}| \textbf{I}_{t}, \textbf{I}_{t-1})$, respectively. These designs ensure that configurations 2 and 4 learn DVF estimation based on warping intensity-based SAX volume data due to $p(\textbf{I}_{t}| \textbf{I}_{t-1}, \textbf{D}_{t})$ that is modelled as  $\hat{\mathbf{I}}_{t} \approx \mathbf{I}_{t-1} \circ \mathbf{D}_{t}$, where $\hat{\mathbf{I}}_{t}$ is the warped SAX volume at time \textit{t}. Consequently, these configurations do not explicitly encourage focus on cardiac anatomical regions. We also model the prior and posterior for configuration 3 by $\mathcal{P} = \prod_{t=0}^{T-1}p(\textbf{I}_{t}| \textbf{I}_{t-1}, \textbf{D}_{t})p(\textbf{M}_{t}| \textbf{M}_{t-1}, \textbf{D}_{t})p(\textbf{D}_{t})$ and $\mathcal{Q} =\prod_{r=0}^{T-1}q(\textbf{M}_{r}| \textbf{I}_{r})\prod_{t=0}^{T-1} q(\textbf{D}_{t}| \textbf{I}_{t}, \textbf{I}_{t-1})$ where $0\leq r \leq T-1, r \notin \{t_{ED}, t_{ES}\}$. The model in configuration 3 learns DVF estimation by matching warped SAX volumes and cardiac shape masks, while the model in configuration 1 learns motion estimation by matching warped cardiac shape masks.

As reported in Tables \ref{table.4} and \ref{table.4mm}, our framework achieves the best cardiac shape registration performance, with the highest DSC and the lowest HD95 across all cardiac anatomical regions when all components are enabled. It indicates that all components contribute to achieving the best performance. These results also show that when \textit{SegNet} is disabled, the cardiac shape registration performance declines more compared to when \textit{RVAE} is disabled. It shows that \textit{SegNet} has a greater impact on the cardiac motion estimation performance of the proposed framework than \textit{RVAE}. Moreover, it shows that the model performs better at cardiac shape registration when it is supposed to focus on cardiac anatomical regions than when it is not. These results also indicate that adding \textit{RVAE} to \textit{DeformNet} without using \textit{SegNet} could slightly improve the cardiac motion estimation performance. Nonetheless, the joint use of \textit{SegNet} and \textit{RVAE} significantly improves cardiac shape registration performance. Consequently, these findings indicate that the \textit{SegNet} module, which enforces our method to supervise DVF estimation through matching cardiac shape masks, contributes most significantly to achieving robust and accurate cardiac shape registration, while the \textit{RVAE} module provides the second-largest contribution. In addition, comparing configurations 1 and 3 indicates that combining intensity-based and shape-matching loss terms reduces cardiac shape registration performance, whereas using only the shape-matching registration loss yields the best performance.

\begin{table}[t!]
  \caption{Quantitative comparison of computing clinical indices on the UK Biobank dataset in terms of the RMSE between the parameters computed from the ground-truth masks at ED and ES time points and those calculated from the cardiac shape masks obtained by the methods at the mentioned time steps. All metrics are reported as Mean$\pm$Std. \textbf{Bold} items indicate the best results.}\label{table.ci_ukb}
  \centering{\LARGE
  \resizebox{1.0\textwidth}{!}{%
    \begin{tabular}{c c c c c c c c c c}
      \hline
      \hline
      Method & LVEDV & LVESV & RVEDV & RVESV & LVSV & RVSV & LVMM & LVEF & RVEF \\
      \hline
      CardioMorphNet & \textbf{21.69$\pm$20.36} & 19.22$\pm$21.78 & \textbf{36.75$\pm$35.42} & \textbf{19.28$\pm$17.38} & \textbf{26.81$\pm$24.93} & \textbf{41.10$\pm$36.48} & 25.85$\pm$21.00 & \textbf{3.10$\pm$3.40} & \textbf{3.93$\pm$3.71} \\
      \hline
      VoxelMorph & 138.62$\pm$57.26 & 115.61$\pm$50.13 & 286.99$\pm$72.93 & 240.94$\pm$74.63 & 254.07$\pm$102.90 & 527.94$\pm$140.06 & \textbf{21.92$\pm$18.35} & 35.77$\pm$14.32 & 101.94$\pm$32.98 \\
      \hline
      TLRN & 134.86$\pm$67.79 & 138.75$\pm$52.90 & 283.34$\pm$72.73 & 289.11$\pm$76.81 & 273.41$\pm$115.15 & 572.45$\pm$143.01 & 30.22$\pm$23.69 & 40.82$\pm$17.34 & 114.17$\pm$35.14 \\
      \hline
      TCSF & 41.25$\pm$29.69 & \textbf{15.18$\pm$18.46} & 102.30$\pm$49.11 & 22.07$\pm$17.56 & 49.77$\pm$37.18 & 120.84$\pm$53.22 & 26.22$\pm$19.74 & 4.51$\pm$3.77 & 11.29$\pm$5.28 \\
      \hline
      SyN & 135.14$\pm$66.35 & 544.27$\pm$117.92 & 510.87$\pm$103.59 & 142.56$\pm$57.00 & 674.88$\pm$159.43 & 651.30$\pm$153.22 & 60.48$\pm$36.79 & 125.13$\pm$29.10 & 262.52$\pm$96.97 \\
      \hline
      SegMorph & 68.52$\pm$48.60 & 90.60$\pm$51.35 & 231.00$\pm$71.68 & 171.72$\pm$63.03 & 156.67$\pm$92.99 & 402.66$\pm$125.83 & 23.02$\pm$21.52 & 21.39$\pm$12.16 & 65.91$\pm$25.01 \\
      \hline
      LCC-Demons & 208.76$\pm$66.84 & 566.52$\pm$118.79 & 520.39$\pm$107.08 & 171.49$\pm$61.35 & 774.10$\pm$172.68 & 689.85$\pm$163.49 & 97.41$\pm$36.25 & 160.09$\pm$32.37 & 308.41$\pm$114.92 \\
      \hline
      DragNet & 110.44$\pm$59.97 & 98.79$\pm$52.29 & 223.39$\pm$65.73 & 167.86$\pm$71.66 & 208.77$\pm$103.74 & 391.25$\pm$124.54 & 22.72$\pm$20.12 & 28.21$\pm$13.86 & 63.17$\pm$25.28 \\
      \hline
      JMS & 116.57$\pm$49.10 & 57.21$\pm$31.24 & 291.35$\pm$71.47 & 33.77$\pm$28.35 & 172.76$\pm$72.31 & 308.07$\pm$90.05 & 67.87$\pm$27.36 & 20.72$\pm$9.62 & 40.51$\pm$18.98 \\
      \hline
      \hline
    \end{tabular}%
    }}
\end{table}

\begin{table}[t!]
  \caption{Quantitative comparison of computing clinical indices on the M\&M dataset in terms of the RMSE between the parameters computed from the ground-truth masks at ED and ES time points and those calculated from the cardiac shape masks obtained by the methods at the mentioned time steps. All metrics are reported as Mean$\pm$Std. \textbf{Bold} items indicate the best results.}\label{table.ci_mm}
  \centering{\LARGE
  \resizebox{1.0\textwidth}{!}{%
    \begin{tabular}{c c c c c c c c c c}
      \hline
      \hline
      Method & LVEDV & LVESV & RVEDV & RVESV & LVSV & RVSV & LVMM & LVEF & RVEF \\
      \hline
       CardioMorphNet
       & \textbf{89.59$\pm$88.03}
       & \textbf{57.39$\pm$52.56}
       & \textbf{97.73$\pm$100.63}
       & \textbf{101.29$\pm$79.05}
       & \textbf{113.95$\pm$94.17}
       & \textbf{133.19$\pm$107.00}
       & \textbf{57.75$\pm$54.23}
       & \textbf{7.53$\pm$5.88}
       & \textbf{12.37$\pm$8.90} \\
      \hline
       VoxelMorph & 475.36$\pm$223.36 & 332.85$\pm$164.62 & 421.75$\pm$198.08 & 291.44$\pm$158.72 & 808.20$\pm$380.02 & 712.38$\pm$350.87 & 85.47$\pm$78.05 & 116.17$\pm$61.74 & 110.47$\pm$62.90 \\
      \hline
       TLRN & 289.16$\pm$157.49 & 199.84$\pm$114.50 & 343.32$\pm$180.10 & 247.70$\pm$135.59 & 488.89$\pm$264.66 & 590.04$\pm$310.87 & 76.27$\pm$65.51 & 50.41$\pm$34.84 & 78.77$\pm$50.66 \\
      \hline
       TCSF & 366.56$\pm$182.12 & 324.21$\pm$160.98 & 431.94$\pm$207.56 & 377.04$\pm$180.39 & 690.77$\pm$338.85 & 808.98$\pm$385.20 & 66.77$\pm$59.88 & 82.72$\pm$51.16 & 134.60$\pm$82.12 \\
      \hline
       SyN & 223.19$\pm$157.60 & 941.17$\pm$348.52 & 664.65$\pm$262.11 & 172.66$\pm$126.68 & 1116.24$\pm$502.30 & 816.91$\pm$396.62 & 117.10$\pm$117.16 & 132.66$\pm$53.66 & 337.71$\pm$1234.71 \\
      \hline
      SegMorph & 250.15$\pm$151.78 & 183.39$\pm$118.76 & 346.50$\pm$204.34 & 347.86$\pm$235.02 & 401.48$\pm$211.78 & 672.93$\pm$330.46 & 116.91$\pm$110.28 & 37.09$\pm$22.23 & 92.98$\pm$44.09 \\
      \hline
      LCC-Demons & 296.94$\pm$174.97 & 1027.49$\pm$382.69 & 684.03$\pm$270.66 & 218.46$\pm$162.03 & 1302.82$\pm$550.26 & 891.51$\pm$426.00 & 155.58$\pm$111.76 & 166.58$\pm$59.84 & 383.61$\pm$964.92 \\
      \hline
      DragNet & 206.88$\pm$156.84 & 209.74$\pm$140.91 & 244.68$\pm$166.05 & 198.05$\pm$147.44 & 415.60$\pm$286.90 & 441.34$\pm$303.14 & 80.91$\pm$75.01 & 40.64$\pm$35.65 & 54.06$\pm$58.33 \\
      \hline
      JMS & 163.52$\pm$180.27 & 130.52$\pm$135.83 & 241.30$\pm$186.42 & 167.21$\pm$144.81 & 173.41$\pm$129.83 & 270.14$\pm$189.82 & 180.60$\pm$169.09 & 14.39$\pm$12.72 & 24.08$\pm$18.81 \\
      \hline
      \hline
    \end{tabular}%
    }}
\end{table}

\subsection{Regularisation Analysis for the Smoothness of DVFs}

The diffeomorphic property is important in medical image registration as NJD values of DVFs indicate non-physical folding and negate the topological plausibility of the estimated DVFs. Consequently, it is required to obtain the optimal balance between cardiac shape registration accuracy and lowest NJD, ensuring that the estimated DVFs are not only accurate with respect to cardiac shape consistency but also smooth with at least number of folding.

\begin{figure}[b!]
    \centering
    \includegraphics[width=0.6\textwidth]{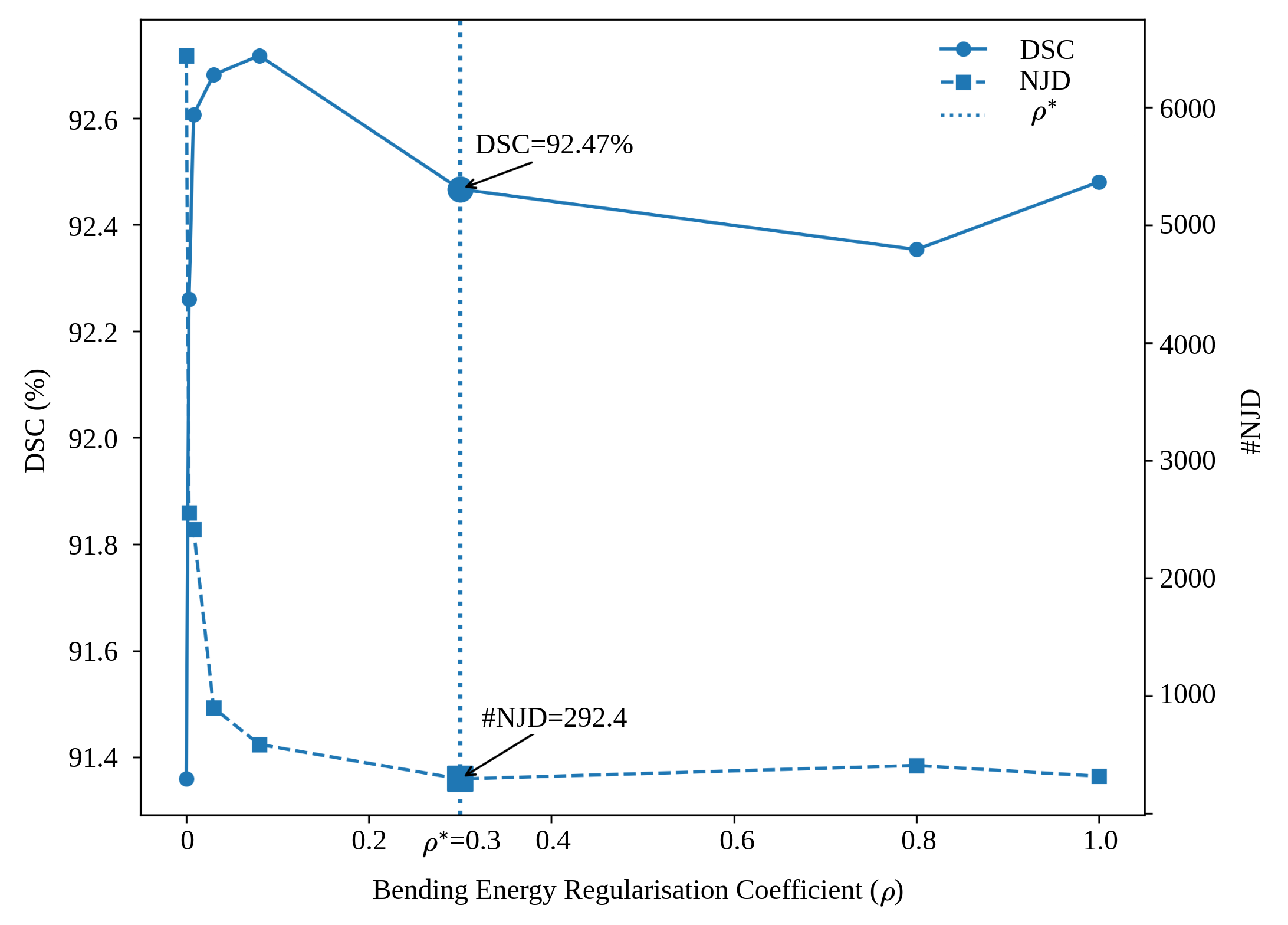}
    \caption{\fontsize{8pt}{10pt}\selectfont Plot of the percentage of mean of DSC and the number of voxels with NJD (\#NJD) obtained on the validation set of UK Biobank dataset across bending energy regularisation coefficients $\rho \in [0, 1.0]$. The plot shows the percentage of DSC and the number of voxels with NJD (\#NJD) obtained on the validation set across bending energy regularisation coefficients $\rho \in [0, 1.0]$. Increasing $\rho$ reduces \#NJD, indicating smoother DVFs, while very large values slightly reduce DSC. The optimal regularisation coefficient $\rho^*$ is selected as 0.3, achieving $\mathrm{DSC}=92.47\%$ and $\mathrm{NJD}=292.4$.}
    \label{fig_dice_vs_ndj}
\end{figure}

Figure \ref{fig_dice_vs_ndj} illustrates the obtained DSC and the number of voxels with NJD by the proposed framework across bending energy regularisation coefficients $\rho \in [0,1]$. The quantitative values of Figure \ref{fig_dice_vs_ndj} are listed in Table \ref{table.lambda_smooth}. As shown in Figure \ref{fig_dice_vs_ndj} and Table \ref{table.lambda_smooth}, increasing the regularisation coefficient from $\rho=0$ to $\rho=0.08$ leads to a gradual improvement in DSC from $91.36\%$ to a maximum of $92.72\%$, suggesting that moderate smoothness constraints encourage more anatomically consistent motion estimation. At the same time, the number of voxels with NJD decreases substantially from $6439.77$ at $\rho=0$ to $585.70$ at $\rho=0.08$, indicating that regularisation effectively suppresses the folding occurrences in the estimated DVFs and results in smoother DVF estimation by the presented framework. When the regularisation weight increases beyond $\rho=0.08$, the DSC shows a slight decline, reaching $92.47\%$ at $\rho=0.3$ and $92.35\%$ at $\rho=0.8$. This behaviour suggests that overly strong smoothness constraints may restrict the flexibility of the registration framework to capture fine-scale anatomical variations in cardiac motion. However, stronger regularisation continues to reduce folding occurrences, achieving the lowest number of voxels with NJD of $292.38$ at $\rho=0.3$, which corresponds to the most balanced DVF estimation between cardiac shape registration accuracy and the smoothness of DVFs.

These results demonstrate a trade-off between cardiac shape registration accuracy and the smoothness of DVFs. While $\rho=0.08$ provides the highest DSC, $\rho=0.3$ achieves the best diffeomorphic property with only a minor reduction in DSC. Therefore, $\rho^{*}=0.3$ is selected as the final bending energy regularisation coefficient, as it provides a suitable balance between accurate cardiac shape matching and the estimated DVFs with the minimum folding occurrences. These findings highlight the importance of selecting an appropriate regularisation coefficient to achieve a compromise between cardiac shape registration accuracy and the smoothness of DVFs in deformable cardiac image registration.

\section{Discussions}\label{sec5}
\begin{table}[t!]
  \caption{Results of the ablation study of the CardioMorphNet framework given the UK Biobank dataset for cardiac motion estimation. \cmark\ and \xmark\ denote the corresponding component is enabled and disabled, respectively. All metrics are reported as Mean$\pm$Std. \textbf{Bold} items indicate the best results.}\label{table.4}
  \centering
  \resizebox{0.9\textwidth}{!}{%
    \begin{tabular}{cccccccccc}
      \hline
      \hline
      \multirow{2}{*}{Config No}  & \multirow{2}{*}{\textit{DeformNet}} &  \multirow{2}{*}{\textit{RVAE}} &  \multirow{2}{*}{\textit{SegNet}} & \multicolumn{3}{c}{DSC(\%)} & \multicolumn{3}{c}{HD95} \\
      \cmidrule{5-10}
      {} & {} & {} & {} & LV & LVmyo & RV & LV & LVmyo & RV \\
      \hline
      1 & \cmark & \cmark & \cmark & \textbf{93.77$\pm$2.11} & \textbf{83.84$\pm$4.20} & \textbf{90.42$\pm$3.20} & \textbf{1.44$\pm$0.65} &  \textbf{1.50$\pm$0.49}&  \textbf{1.50$\pm$0.88} \\
      \hline
        2 & \cmark & \cmark & \xmark & 80.42$\pm$2.84 & 59.16$\pm$7.10 & 78.49$\pm$3.47 & 4.67$\pm$2.74 & 3.96$\pm$1.42 & 5.37$\pm$4.57 \\
      \hline
        3 & \cmark & \xmark & \cmark & 87.00$\pm$4.17 & 73.57$\pm$6.00 & 82.28$\pm$4.79 & 2.42$\pm$1.25 & 2.17$\pm$0.62 & 2.93$\pm$1.81 \\
      \hline
        4 & \cmark & \xmark & \xmark & 78.04$\pm$3.27 & 53.91$\pm$7.95 & 75.53$\pm$4.04 & 4.90$\pm$1.90 & 4.05$\pm$1.02 & 6.50$\pm$5.22 \\
      \hline
      \hline
    \end{tabular}%
    }
\end{table}

\begin{table}[t!]
  \caption{Results of the ablation study of the CardioMorphNet framework given the M\&M dataset for cardiac motion estimation. \cmark\ and \xmark\ denote the corresponding component is enabled and disabled, respectively. All metrics are reported as Mean$\pm$Std. \textbf{Bold} items indicate the best results.}\label{table.4mm}
  \centering
  \resizebox{0.9\textwidth}{!}{%
    \begin{tabular}{cccccccccc}
      \hline
      \hline
      \multirow{2}{*}{Config No}  & \multirow{2}{*}{\textit{DeformNet}} &  \multirow{2}{*}{\textit{RVAE}} &  \multirow{2}{*}{\textit{SegNet}} & \multicolumn{3}{c}{DSC(\%)} & \multicolumn{3}{c}{HD95} \\
      \cmidrule{5-10}
      {} & {} & {} & {} & LV & LVmyo & RV & LV & LVmyo & RV \\
      \hline
      1 & \cmark & \cmark & \cmark & \textbf{89.99$\pm$4.18} & \textbf{80.36$\pm$4.56} & \textbf{84.94$\pm$6.92} & \textbf{2.22$\pm$2.08} &  \textbf{1.98$\pm$0.75}&  \textbf{2.53$\pm$2.84} \\
      \hline
        2 & \cmark & \cmark & \xmark & 74.15$\pm$10.18 & 56.80$\pm$9.62 & 64.77$\pm$10.85 & 6.11$\pm$3.42 & 4.93$\pm$1.97 & 9.21$\pm$6.62 \\
      \hline
        3 & \cmark & \xmark & \cmark & 77.59$\pm$6.22 & 58.58$\pm$8.44 & 74.46$\pm$6.06 & 5.87$\pm$5.01 & 4.16$\pm$2.20 & 5.50$\pm$5.27 \\
      \hline
        4 & \cmark & \xmark & \xmark & 60.55$\pm$13.15 & 41.34$\pm$11.69 & 59.39$\pm$9.94 & 10.61$\pm$4.48 & 6.11$\pm$1.58 & 10.71$\pm$6.64 \\
      \hline
      \hline
    \end{tabular}%
    }
\end{table}

In this study, we introduce a 3D probabilistic sequential cardiac shape-guided registration framework based on cine CMR SAX volumes throughout the cardiac cycle. Our probabilistic modelling using variational Bayes allows us to derive precise DVFs from two sequential SAX volumes and to focus on the cardiac anatomical regions for motion estimation by minimising the KL divergence between the distributions of the segmentation component outputs of CardioMorphNet (\textit{SegNet}) and the distributions of the warped segmentation maps. The framework further benefits from an \textit{RVAE} model that captures the distribution of spatio-temporal dependencies throughout the cardiac cycle, as well as from explicit modelling of the mean and covariance of the DVFs, enabling quantification of DVF uncertainties without utilising dropout-based methods. The 3D modelling also allows us to derive 3D DVFs for cardiac motion estimation.

In the presented framework, the DVFs are modelled retroactively using the information from the previous cardiac time points through formulating the joint distribution $\mathcal{P} = p(\mathcal{M}, \mathcal{I}, \mathcal{D}, \mathcal{Z})$, which is provided in Eq. \eqref{eq1}. To capture this spatio-temporal information over the cardiac cycle, a recurrent latent variable is modelled in this framework. This latent variable is learned from the current SAX volume and the hidden state, which is updated recursively using a ConvLSTM network. The ConvLSTM network employs the features of the current SAX data, the latent variable, and the previous hidden state to capture temporal variations across the SAX sequence. We condition the current mask volume on the previous mask and DVF, $p(\textbf{M}_{t}| \textbf{M}_{t-1}, \textbf{D}_{t})$, in the joint distribution, and this variable is conditioned on the current SAX volume in the posterior distribution, $q(\textbf{M}_{t}|\textbf{I}_{t})$. This modelling enables the framework to warp the cardiac shape masks, provided by the posterior function, over the cardiac cycle using the estimated DVFs. On the other hand, the DVF at each time step is estimated using $q(\textbf{D}_{t}| \textbf{I}_{t}, \textbf{I}_{t-1}, \textbf{Z}_{t})$ given two sequential SAX volumes and the latent variable. By minimising the ELBO loss in Eq. \eqref{eq6}, the framework learns to extract multi-scale features from the paired SAX volumes, utilise spatio-temporal dependencies in the latent variable, and focus on cardiac anatomical regions for DVF estimation. This anatomical focus is achieved by minimising the supervised and semi-supervised shape loss terms, which are provided in Eqs. \eqref{eq7} and \eqref{eq8}, allowing the framework to be guided by the cardiac shape information at each cardiac time step. Our Bayesian modelling also prevents the proposed registration model from being directly trained based on the intensity-based image registration similarity loss between the moved and fixed SAX volumes, which may result in capturing various DVFs that are unrelated to cardiac motion. In terms of implementation, the segmentation component of CardioMorphNet, which acts as $q(\textbf{M}_{t}| \textbf{I}_{t})$, is pre-trained to learn cardiac anatomical regions segmentation in SAX volumes. This enables the registration branch of our framework to be guided by the segmentation component during the second training phase, leading the framework to learn cardiac motion estimation while accounting for cardiac shape consistency.

\begin{table}[b!]
    \caption{Quantitative results of the proposed framework for cardiac shape registration and diffeomorphic property in terms of the percentage (\%) of DSC and the number of voxels with NJD (\#NJD) across bending energy regularisation coefficients $\rho \in [0, 1.0]$. The results obtained using the validation set of UK Biobank dataset. DSC is computed for ED and ES time points between cardiac shape ground-truth masks and the moved shape masks by the proposed framework. All metrics are reported as Mean$\pm$Std. \textbf{Bold} items indicate the best results.}
    \label{table.lambda_smooth}
    \centering
    \resizebox{0.3\textwidth}{!}{%
    \begin{tabular}{ccc}
        \hline
        \hline
        $\rho$ & DSC(\%) & \#NJD \\
        \hline
        0.000 & 91.36 $\pm$ 0.48 & 6439.77 $\pm$ 955.77 \\
        0.003 & 92.26 $\pm$ 0.49 & 2557.57 $\pm$ 189.29 \\
        0.008 & 92.61 $\pm$ 0.44 & 2412.67 $\pm$ 265.32 \\
        0.030 & 92.68 $\pm$ 0.44 & 895.47 $\pm$ 102.40 \\
        0.080 & \textbf{92.72 $\pm$ 0.45} & 585.70 $\pm$ 73.97 \\
        0.300 & 92.47 $\pm$ 0.52 & \textbf{292.38 $\pm$ 54.61} \\
        0.800 & 92.35 $\pm$ 0.51 & 406.99 $\pm$ 71.97 \\
        1.000 & 92.48 $\pm$ 0.51 & 315.82 $\pm$ 74.98 \\
        \hline
        \hline
    \end{tabular}%
    }
\end{table}

Based on the reported results, the proposed registration framework achieves significantly higher DSC and JAC values and lower HD95 and MSD metrics for cardiac shape registration across both the UK Biobank and M\&M datasets, compared to other state-of-the-art methods. These results indicate that our framework outperforms other approaches for cardiac motion estimation. The main reason for this superiority is that the presented registration framework is guided by cardiac shape information, while leveraging features of two sequential SAX volumes and a latent variable representing spatio-temporal features over the cardiac cycle. It leads the framework to extract and utilise multi-scale spatial and temporal features over the cardiac cycle and focus on the cardiac anatomical regions for cardiac motion estimation. At the same time, traditional methods perform registration using intensity-based similarity metrics between the fixed and warped images, which limits their ability to focus on regions containing discontinuities, such as tissue boundaries and anatomical structures, and results in the capture of various motions irrelevant to the heart.

Regarding diffeomorphic properties, our method does not explicitly enforce global diffeomorphic constraints on the estimated DVFs. Instead, the DVFs are supervised through alignment of discrete cardiac shape masks, which prioritises anatomically consistent shape matching rather than strict global smoothness. Consequently, the proposed method exhibits slightly higher NJD values than approaches that impose strong global smoothness constraints, such as intensity-driven registration methods. This reflects an inherent trade-off between achieving precise anatomical boundary alignment and enforcing strict diffeomorphic regularisation. Importantly, this behaviour enables the framework to focus on anatomically meaningful cardiac regions during DVF estimation, rather than enforcing uniform smoothness in regions that are not directly relevant to cardiac structures. Similar observations have been reported in prior studies \cite{chen2022joint, pace2013locally, hua2017multiresolution}, which show that cardiac motion fields may exhibit locally non-smooth or discontinuous behaviour near anatomical boundaries. In contrast, methods that rely on intensity-based similarity losses with strong smoothness regularisation tend to estimate overly smooth DVFs, which may overlook cardiac anatomical structures during DVF estimation and capture motion patterns that are less consistent with cardiac structural motion. However, excessive non-smooth DVF estimation increases the number of folding of DVFs and could suppress the physical plausibility of estimated motions. Hence, it is essential to obtain the optimal balance of DVF estimation between anatomical shape consistency registration and the diffeomorphic property of DVFs by choosing the optimal bending-energy regularisation coefficient. According to the obtained results, this parameter significantly influences the diffeomorphic property and the number of voxels with NJD, which reduces the physical plausibility of DVFs, while having a comparatively smaller effect on DSC for cardiac shape registration. Therefore, it should be selected to achieve a balanced motion estimation performance between diffeomorphic behaviour and cardiac shape registration accuracy.

Compared with joint segmentation-registration techniques such as SegMorph, JMS, and TCSF approaches that aim to provide pseudo-ground-truth masks for the segmentation network branch via image registration to improve segmentation, our framework enforces the registration branch to concentrate on the cardiac anatomical regions through matching the warped cardiac shape segmentation maps and the fixed ones that are provided by a segmentation network. It leads to our method producing cardiac shape masks that are better aligned with cardiac anatomical regions than those produced by the SegMorph, JMS, and TCSF methods. Compared with other Bayesian-based approaches, DragNet and SegMorph, the DVFs estimated by the proposed framework are less uncertain in the cardiac anatomical regions, as evidenced in Figures \ref{fig_uncmaps} and \ref{fig_boxplot_unc}. This is another advantage of our formulation, which supervises DVF estimation based on anatomical shape consistency rather than intensity-based image registration similarity loss, as used in DragNet and SegMorph. As a disadvantage, our method is unable to generate realistic cine SAX sequences using the \textit{RVAE} component, whereas DragNet and SegMorph can perform this task by conditioning the posterior of DVFs on only the moving image and the latent variable.

In terms of estimating clinical indices, the proposed framework outperforms other cardiac motion estimation approaches, including conventional registration methods and joint segmentation--registration techniques, by yielding values closer to the reference values. Although the clinical indices evaluated in this study are static quantities that can also be derived using segmentation networks, the superior performance of the proposed method compared with other registration-based approaches suggests an improved capability to estimate motion-related biomarkers. In particular, myocardial strain quantification that reflects the relative deformation of the myocardium along three orthogonal directions, such as radial, circumferential, and longitudinal, throughout the cardiac cycle \cite{sinclair2018myocardial, ferdian2020fully} can be derived using motion estimation approaches. This index is an important dynamic indicator for early diagnosis of myocardial ischemia \cite{smiseth2025myocardial, diao2017diagnostic}. Nonetheless, direct evaluation of myocardial strain was not feasible in this study because tagged MRI data or reliable ground-truth strain measurements were not available in the datasets used. Therefore, static clinical indices were used as an indirect approach to assess the performance of cardiac motion estimation relative to other methods. In future work, we will consider dynamic clinical indices, such as myocardial strain, to enable a more comprehensive evaluation of cardiac motion estimation.

Although the proposed framework improves cardiac motion estimation using cine SAX volumes compared to state-of-the-art methods, it suffers from several limitations. The first limitation is the high computational load of the model during training, which limits the selection of the number of SAX frames. Another restriction is segmentation errors in cardiac shape masks introduced by the segmentation module, which may affect cardiac shape registration performance. Moreover, there is no correspondence between the estimated motions in each voxel across various subjects, restricting intra-subject cardiac motion assessment. This is particularly important because many pathological abnormal motions are defined relative to a population distribution of anatomically corresponding points/voxels. To address this, our future work will focus on cardiac motion estimation in the mesh domain using the cardiac atlas meshes provided by \cite{bai2015bi} to predict cardiac motion with globally corresponding points across various patients.

\section{Conclusion} \label{sec6}
This study introduces a shape-guided probabilistic recurrent deep learning model for cardiac motion estimation using cine CMR SAX data throughout the cardiac cycle. The formulation of our method enables us to derive DVFs from two sequential SAX volumes and to leverage spatio-temporal dependencies provided by an \textit{RVAE} model, while focusing on the cardiac region for DVF estimation by minimising the KL divergence between the warped and fixed cardiac shapes, as provided by the segmentation branch of the presented framework. Unlike traditional methods that rely on intensity-based image registration similarity loss, our approach is independent of this term. Instead, it aligns warped and fixed segmentation maps by incorporating features from sequential SAX volumes and recurrent spatio-temporal dependencies. According to the obtained results, the proposed framework achieves superior performance in cardiac shape registration compared to other state-of-the-art methods. Moreover, the DVFs obtained by the proposed framework are more confident in the cardiac region than those from other probabilistic methods, such as DragNet and SegMorph, as evidenced by the uncertainty assessment. The results also show that the presented method estimates clinical indices more accurately than other approaches, which could help cardiologists identify early CVDs and cardiac motion abnormalities. While our method outperforms other approaches for cardiac motion estimation, it still suffers from segmentation errors that may affect registration performance and a high computational load during the training phase, limiting the number of input SAX frames throughout the cardiac cycle. It is also unable to estimate cardiac motion in the global corresponding points/voxels across various subjects, restricting intra-subject cardiac motion assessment.

\appendix

\section{Detailed Network Architecture}
\label{app:architecture}
\begin{figure}[t!]
    \centering
    \includegraphics[width=0.8\textwidth]{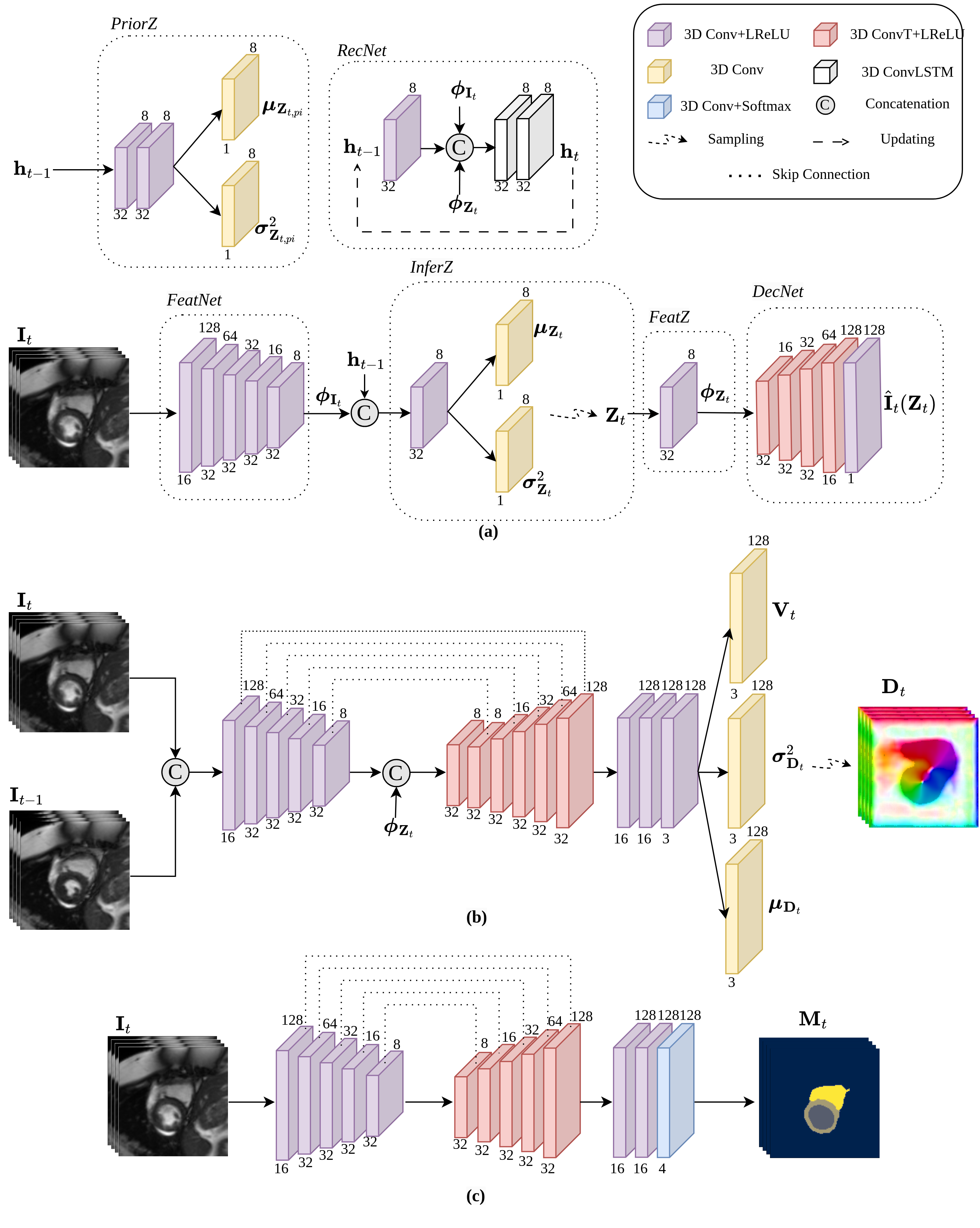}
    \caption{\fontsize{8pt}{10pt}\selectfont
    Detailed deep learning architecture of the proposed \textit{CardioMorphNet} framework, consisting of three main components:
    (\textbf{a}) the \textit{RVAE} module for spatio-temporal latent representation learning,
    (\textbf{b}) the \textit{DeformNet} module for predicting the DVF, and
    (\textbf{c}) the \textit{SegNet} module for outputting cardiac shape masks.
    For each convolutional block, the upper number indicates the spatial resolution (height and width) of the feature map, while the lower number denotes the number of feature channels.
    All convolutional layers employ a unit stride along the depth dimension to preserve inter-slice resolution.
    The \textit{RVAE} component integrates a variational autoencoder with a recurrent ConvLSTM (\textit{RecNet}) to model spatio-temporal dependencies using the latent variable $\mathbf{Z}_{t}$.
    \textit{DeformNet} is a U-Net model that leverages the concatenation of $\mathbf{I}_{t}$ and $\mathbf{I}_{t-1}$ as the primary input and incorporates latent features $\boldsymbol{\phi}_{\mathbf{Z}_{t}}$ at the bottleneck to estimate both the mean and variance of the DVF.
    \textit{SegNet} is another U-Net model that outputs cardiac shape masks.}
    \label{fig_deep_arc_digram}
\end{figure}

The deep learning network architecture of the proposed framework is shown in Figure \ref{fig_deep_arc_digram}. As shown in Figure \ref{fig_deep_arc_digram}, our framework is structured by 3D convolution layers (3D Conv) for feature extraction and downsampling in the height and width dimensions, and 3D transposed convolution layers (3D ConvT) for upscaling and reconstruction. The strides of these layers are set to 2, 2, and 1 along the height, width, and depth dimensions, respectively. The reason for setting stride 1 along the depth dimension is to preserve it, since SAX data are not high-resolution in this dimension and often have a small number of slices. The Leaky Rectified Linear Unit (LReLU) is also used as the activation layer in our framework to introduce nonlinearity, enabling the model to learn complex patterns and support backpropagation. The segmentation component, \textit{SegNet}, utilises Softmax as its final activation layer to compute the segmentation probability map from the SAX volume.

\section{Comparison with Segmentation Models for Clinical Indices Extraction}
\label{app:seg_results}

Tables \ref{table.ci_ukb_seg} and \ref{table.ci_mm_seg} list the RMSEs of estimated clinical indices obtained by the proposed framework with two segmentation baselines: U-Net and nnU-Net. As expected, these segmentation models achieve very strong performance in estimating static clinical indices, since they predict the ED and ES masks directly. Our method achieves comparable performance while being designed primarily for cardiac motion estimation rather than static segmentation. This comparison clarifies that, although segmentation methods are more natural baselines for static indices, the proposed method remains competitive on these measures while also providing DVFs across the cardiac cycle, which could provide an accurate and robust representation of cardiac motion, enabling the extraction of motion-related downstream tasks, i.e., myocardial strain quantification.

\section*{Acknowledgements}
This study has been done using the UK Biobank Resource under application number 71392. The authors sincerely thank Dr Rona Strawbridge from the School of Health and Wellbeing, University of Glasgow, Glasgow, UK, for granting access to the mentioned application number of the UK Biobank dataset and for her valuable guidance on data usage and interpretation.

\section*{Data statement}
The authors do not have permission to share data.

\section*{Ethics statement}
This study accurately presents the research and methods, including only original work or properly cited material, and is not currently being considered for publication in another journal, with all contributors properly credited. Ethical approval for UK Biobank was obtained from the North West -- Haydock Research Ethics Committee (approval letter dated 29th June 2021, REC reference: 21/NW/0157).

\begin{table}[t!]
  \caption{Quantitative comparison of computing clinical indices on the UK Biobank dataset in terms of the RMSE between the parameters computed from the ground-truth masks at ED and ES time points and those calculated from the cardiac shape masks obtained by the methods at the mentioned time steps. All metrics are reported as Mean$\pm$Std. \textbf{Bold} items indicate the best results.}\label{table.ci_ukb_seg}
  \centering{\LARGE
  \resizebox{1.0\textwidth}{!}{%
    \begin{tabular}{c c c c c c c c c c}
      \hline
      \hline
      Method & LVEDV & LVESV & RVEDV & RVESV & LVSV & RVSV & LVMM & LVEF & RVEF \\
      \hline
      CardioMorphNet & 21.69$\pm$20.36 & 19.22$\pm$21.78 & 36.75$\pm$35.42 & 19.28$\pm$17.38 & 26.81$\pm$24.93 & 41.10$\pm$36.48 & 25.85$\pm$21.00 & 3.10$\pm$3.40 & 3.93$\pm$3.71 \\
      \hline
      U-Net & 15.66$\pm$16.65 & 16.80$\pm$19.41 & 26.55$\pm$28.94 & 18.66$\pm$17.38 & 21.04$\pm$19.81 & 31.99$\pm$29.66 & 16.81$\pm$15.00 & 2.62$\pm$2.76 & 2.99$\pm$2.99 \\
      \hline
      nnU-Net & \textbf{10.86$\pm$13.32} & \textbf{9.39$\pm$13.87} & \textbf{17.67$\pm$17.08} & \textbf{10.13$\pm$9.87} & \textbf{13.96$\pm$17.94} & \textbf{19.56$\pm$18.37} & \textbf{9.71$\pm$8.03} & \textbf{1.55$\pm$2.10} & \textbf{1.85$\pm$1.75} \\
      \hline
      \hline
    \end{tabular}%
    }}
\end{table}

\begin{table}[t!]
  \caption{Quantitative comparison of computing clinical indices on the M\&M dataset in terms of the RMSE between the parameters computed from the ground-truth masks at ED and ES time points and those calculated from the cardiac shape masks obtained by the methods at the mentioned time steps. All metrics are reported as Mean$\pm$Std. \textbf{Bold} items indicate the best results.}\label{table.ci_mm_seg}
  \centering{\LARGE
  \resizebox{1.0\textwidth}{!}{%
    \begin{tabular}{c c c c c c c c c c}
      \hline
      \hline
      Method & LVEDV & LVESV & RVEDV & RVESV & LVSV & RVSV & LVMM & LVEF & RVEF \\
      \hline
       CardioMorphNet
       & 89.59$\pm$88.03
       & 57.39$\pm$52.56
       & 97.73$\pm$100.63
       & 101.29$\pm$79.05
       & 113.95$\pm$94.17
       & 133.19$\pm$107.00
       & \textbf{57.75$\pm$54.23}
       & 7.53$\pm$5.88
       & 12.37$\pm$8.90 \\
      \hline
      U-Net & \textbf{65.90$\pm$74.91} & \textbf{41.84$\pm$38.46} & \textbf{89.03$\pm$101.77} & \textbf{68.82$\pm$63.92} & \textbf{59.88$\pm$60.91} & 84.91$\pm$89.95 & 81.44$\pm$60.04 & \textbf{3.82$\pm$3.67} & \textbf{6.88$\pm$6.07} \\
      \hline
      nnU-Net & 72.51$\pm$70.24 & 54.66$\pm$49.59 & 99.37$\pm$83.26 & 77.07$\pm$76.64 & 73.39$\pm$64.91 & \textbf{82.64$\pm$68.22} & 115.02$\pm$78.21 & 5.26$\pm$4.75 & 7.46$\pm$8.77 \\
      \hline
      \hline
    \end{tabular}%
    }}
\end{table}

\end{document}